\documentclass[a4paper,fleqn]{cas-dc}
\usepackage[square, numbers]{natbib}

\usepackage{caption}
\usepackage{cas-common}
\usepackage{graphicx}
\usepackage{diagbox}
\usepackage{mathtools}
\usepackage{amsmath}
\usepackage{bbding}
\usepackage{pifont}
\usepackage{wasysym}
\usepackage{tikz}
\usepackage{calc}
\usepackage{multirow}
\usepackage{rotating}
\usepackage{colortbl}
\usepackage{color}
\usepackage{ragged2e}
\usepackage{float}
\usepackage{tabularx} % 放在导言区
\usepackage{booktabs} % 更美观的线条（可选）
\UseRawInputEncoding

\begin{document}
\captionsetup[figure]{labelfont={bf},labelformat={default},name={Fig.}, labelsep=period,}
\captionsetup[table]{labelfont={bf},labelformat={default},labelsep=newline,name={Table}, singlelinecheck=false}
\let\WriteBookmarks\relax
\def\floatpagepagefraction{1}
\def\textpagefraction{.001}

% Short title
\shorttitle{A Comprehensive Survey for Real-World Industrial Defect Detection: Challenge, Approach, and Prospect}    

% Short author
\shortauthors{Y. Cheng et~al. }  

% Main title of the paper
\title [mode = title]{A Comprehensive Survey for Real-World Industrial Defect Detection: Challenges, Approaches, and Prospects}  

% Title footnote mark
% \tnotemark[1]  %%% star on the title
% \tnotemark[<tnote number>] 

% Title footnote 1.
% eg: \tnotetext[1]{Title footnote text}
% \tnotetext[1]{This work was partially supported by the National Natural Science Foundation of China (Grant U2013203, 61973106). Professor Ajmal Mian is the recipient of an Australian Research Council Future Fellowship Award (project number FT210100268) funded by the Australian Government.} 

% First author
%
% Options: Use if required
% eg: \author[1,3]{Author Name}[type=editor,
%       style=chinese,
%       auid=000,
%       bioid=1,
%       prefix=Sir,
%       orcid=0000-0000-0000-0000,
%       facebook=<facebook id>,
%       twitter=<twitter id>,
%       linkedin=<linkedin id>,
%       gplus=<gplus id>]

% \author[<aff no>]{Yongshan~Yu}[<options>]

\author[1]{Yuqi Cheng}[type=editor, auid=000, bioid=1, orcid= 0000-0003-1994-7301]
\cormark[1]
\ead{yuqicheng@hust.edu.cn}

\author[2]{Yunkang Cao}[type=editor,auid=000,bioid=1,orcid= 0000-0001-7619-6618]
\cormark[1]
\ead{caoyunkang@ieee.org}

\author[3]{Haiming Yao}[type=editor,auid=000,bioid=1,orcid=0000-0003-1419-5489]
\ead{yhm22@mails.tsinghua.edu.cn}

\author[3]{Wei Luo}[type=editor,auid=000,bioid=1,orcid=0000-0003-3125-054X]
\ead{luow23@mails.tsinghua.edu.cn}

\author[4]{Cheng Jiang}[type=editor,auid=000,bioid=1,orcid=0000-0003-3084-1321]
\ead{307080666@qq.com}

\author[2]{Hui Zhang}[type=editor,auid=000,bioid=1,orcid= 0000-0002-1803-3148]
\ead{zhanghuihby@126.com}

\author[1]{Weiming Shen}[type=editor,auid=000,bioid=1,orcid= 0000-0001-5204-7992]
\cormark[2]
\ead{wshen@ieee.org}

\cortext[1]{The two authors contribute equally to this work}
\cortext[2]{Corresponding author}

\affiliation[1]{
            organization={State Key Laboratory of Intelligent Manufacturing Equipment and Technology, Huazhong University of Science and Technology},
            % addressline={}, 
            city={Wuhan},
            postcode={430074}, 
            % state={Hunan},
            country={China}}

\affiliation[2]{organization={School of Artificial Intelligence and Robotics, Hunan University},
            % addressline={Acton}, 
            city={Hunan},
            postcode={410082}, 
            % state={ACT},
            country={China}}

\affiliation[3]{organization={Department of Precision Instrument, Tsinghua University},
            % addressline={Acton}, 
            city={Beijing},
            postcode={100084}, 
            % state={ACT},
            country={China}}

\affiliation[4]{organization={Geely Automobile Research Institute (Ningbo) Co., Ltd.},
            % addressline={35 Stirling Hwy}, 
            city={Ningbo},
            postcode={315336}, 
            % state={WA},
            country={China}}
            
% \footnotetext[1]{\dag\ Equal contribution}
% \footnotetext[2]{\ddag\ Corresponding author}

% Here goes the abstract
\begin{abstract}
% Industrial defect detection (DD) is critical for maintaining product quality across modern manufacturing lines.  With increasing demands for precision, automation, and scalability, traditional inspection techniques struggle to meet the requirements in real-world scenarios. Recent advances in computer vision and deep learning have significantly enhanced defect detection capabilities in both 2D and 3D modalities, enabling improved performance in texture and geometry analysis.  In particular, the transition from closed-set to open-set DD has gained substantial traction, reducing reliance on exhaustive defect annotations and enabling the detection of previously unseen anomalies.  Despite these advancements, a unified and up-to-date understanding of industrial DD remains lacking. Therefore, this survey offers a comprehensive review for both close-set and open-set DD methods in either 2D and 3D modalities, showing the scheme transition in recent years and a growing focus on open-set methods. We summarize key challenges in real-world detection, and highlight emerging trends, offering a timely and holistic perspective on this rapidly evolving field.
Industrial defect detection is vital for upholding product quality across contemporary manufacturing systems. As the expectations for precision, automation, and scalability intensify, conventional inspection approaches are increasingly found wanting in addressing real-world demands. Notable progress in computer vision and deep learning has substantially bolstered defect detection capabilities across both 2D and 3D modalities. A significant development has been the pivot from closed-set to open-set defect detection frameworks, which diminishes the necessity for extensive defect annotations and facilitates the recognition of novel anomalies. Despite such strides, a cohesive and contemporary understanding of industrial defect detection remains elusive. Consequently, this survey delivers an in-depth analysis of both closed-set and open-set defect detection strategies within 2D and 3D modalities, charting their evolution in recent years and underscoring the rising prominence of open-set techniques. We distill critical challenges inherent in practical detection environments and illuminate emerging trends, thereby providing a current and comprehensive vista of this swiftly progressing field.
\end{abstract}


\maketitle

% Main text

\section{Introduction}

Industrial inspection plays a pivotal role in ensuring product quality across a variety of manufacturing domains~\citep{xie2024iad,gao2022review}. The ongoing shift toward high-precision and mass-customized production has intensified the demand for defect detection (DD) systems that exhibit accuracy, efficiency, and adaptability to large-scale industrial operations~\citep{cao2024survey}. Conventional techniques, such as manual visual inspections and mechanical tools, are increasingly outdated, constrained by their dependence on human labor and sluggish processing speeds, rendering them incompatible with the automated, high-throughput requirements of modern industry.

The advent of image processing has fundamentally reshaped automated inspection practices~\citep{tao_unsupervised_2022}. Methods such as binary segmentation and template matching have emerged as standard approaches for surface defect detection. Nevertheless, these techniques demand precisely controlled imaging environments and remain sensitive to variations in lighting, object orientation, and shape~\citep{MVTec-AD}. In real-world industrial contexts, DD is further complicated by the intricate geometries and diverse material compositions of components, necessitating bespoke imaging solutions that balance quality with cost-effectiveness~\citep{real-iad,M2AD}. Traditional methods often prove inadequate for these challenges and are typically evaluated under idealized rather than practical conditions~\citep{MVTec-AD,VisA}.

\begin{figure*}[h!]
\centering\includegraphics[width=\linewidth]{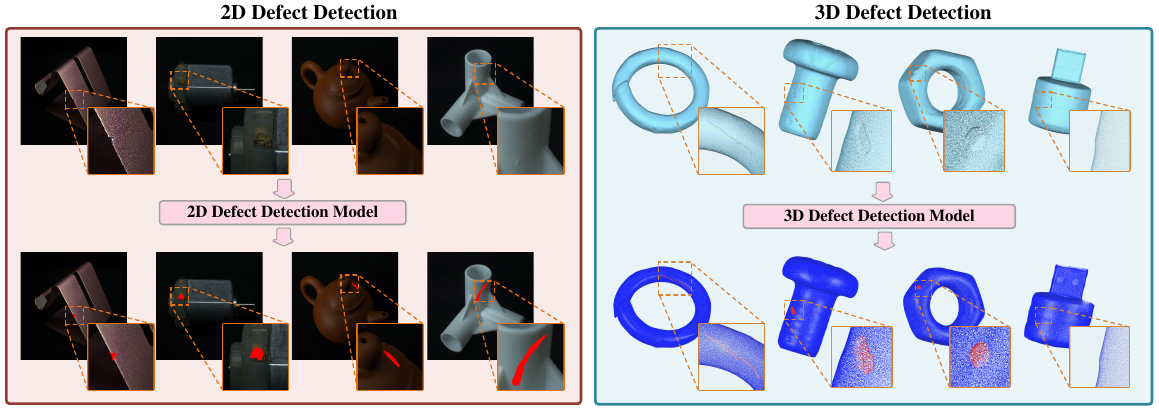}
\caption{\textbf{Illustration of 2D and 3D defect detection paradigms.} \textbf{Left:} 2D detection operates on RGB images to localize surface defects, exemplified using samples from $M^{2}$AD~\cite{M2AD}. \textbf{Right:} 3D detection leverages geometric representations such as point clouds, with data from MiniShift~\cite{Simple3D}. Both approaches aim for precise defect localization, as visualized in the bottom row.}
\vspace{-3mm}
\label{fig:defects}
\end{figure*}

Defects may manifest in both textural and geometric forms, as depicted in Fig. ~\ref{fig:defects}, thereby prompting the exploration of 2D and 3D DD methodologies. 2D detection, utilizing industrial cameras, excels at capturing textural irregularities, whereas 3D detection, employing laser scanners or depth sensors, provides precise geometric detail. Collectively, these approaches facilitate the identification of a broad spectrum of defects. In recent years, both modalities have transitioned from closed-set to open-set detection frameworks. Traditional DD techniques predominantly rely on sophisticated supervised learning methods, such as object detection and segmentation, which require extensive annotated datasets for training. Such datasets are both costly and scarce within industrial settings, limiting these methods to recognizing only known defect categories (closed-set DD). By contrast, open-set DD, often termed industrial visual anomaly detection, seeks to identify novel defects without reliance on annotated anomaly data, thereby catalyzing significant advancements in the field.

Despite the proficiency of various techniques in specific applications, a cohesive understanding of the field remains elusive. Existing reviews~\citep{tao2022deep,lin2025survey,liu2025anomaly,xie2024iad,diers2023survey,gao2022review} frequently confine their focus to specific sub-fields, such as open-set DD or particular anomaly detection strategies, thereby overlooking recent developments in industrial DD. To address this gap, this survey delivers a comprehensive and contemporary analysis of DD methodologies tailored to complex industrial components. To the best of our knowledge, it is the first to systematically examine the transition from closed-set to open-set DD, synthesizing insights across both 2D and 3D modalities. We pledge to periodically update this work to mirror the field's swift evolution.\footnote{Readers are invited to submit relevant updates to yuqicheng@hust.edu.cn.}

\begin{figure*}[h!]
\centering\includegraphics[width=\linewidth]{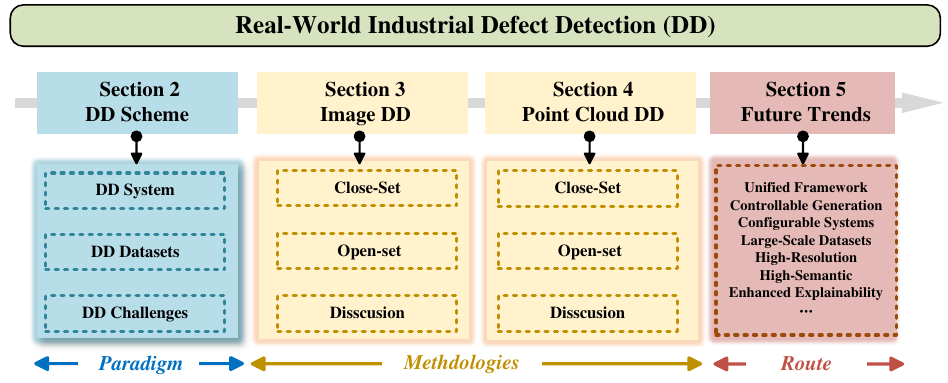}
\caption{{Organization of this survey.}}
\vspace{-3mm}
\label{fig:paper_organization}
\end{figure*}

% Outlining the survey's structure
This survey is structured as follows: Section~\ref{sec:Scheme} elucidates the general DD framework, encompassing typical configurations and real-world challenges. Sections~\ref{sec:image_defect_detection} and~\ref{sec:point_cloud_defect_detection} offer detailed reviews of recent progress in 2D and 3D DD, respectively. Section~\ref{sec:trend} explores emerging trends and prospective research directions. The survey concludes with a summary of key insights in Section~\ref{sec:conclusion}. An illustration of this structure is provided in Fig.~\ref{fig:paper_organization}.

\section{Defect Detection Scheme}\label{sec:Scheme}

Industrial components are characterized by complex geometries and exacting dimensional tolerances, presenting formidable challenges for DD. Contemporary research has progressively delineated a robust detection framework aimed at achieving both efficiency and precision in DD. This section offers a comprehensive review, systematically exploring three critical dimensions: the detection system, datasets, and associated challenges.

\begin{table}[htbp]
\centering
\caption{Mobile platforms and their typical applications.}
\renewcommand{\arraystretch}{1.8} % 调整行距
\resizebox{\linewidth}{!}{
\begin{tabular}{ >{\centering\arraybackslash}m{2.0cm} 
                | >{\centering\arraybackslash}m{1.2cm} 
                | >{\raggedright\arraybackslash}m{7cm} }
\toprule[1.5pt]
Mobile Platform   & Freedom  & \makecell[c]{\centering Typical applications} \\
\hline
Conveyor Belt         & 1 & High-speed inspection of mass-produced items, such as electronic components and packaging, suitable for large-scale manufacturing environments. \\ \hline

Rail System        &  1-3  &  Precision inspection of large-sized components, such as high-precision parts in the automotive and aerospace industries, requiring accurate positioning and measurement. \\ \hline

Turntable         & 1 & Comprehensive surface inspection of components, such as automotive tires and engine housings, enabling 360-degree evaluation of defects and quality.\\ \hline

Robotic Arm         & 3-6 & Precision inspection of complex components with varying shapes, angles, and positioning requirements, suitable for dynamic and versatile quality control tasks. \\
\toprule[1.5pt]
\end{tabular}
}
\label{platform}
\end{table}
\begin{table}[htbp]
\centering
\caption{Sensors and their typical applications.}
\renewcommand{\arraystretch}{1.8} % 调整行距
\resizebox{\linewidth}{!}{
\begin{tabular}{ >{\centering\arraybackslash}m{2.0cm} 
                | >{\centering\arraybackslash}m{1.2cm} 
                | >{\raggedright\arraybackslash}m{7cm} }
\toprule[1.5pt]
Sensor & Modality & \makecell[c]{\centering Typical applications} \\
\hline
RGB Camera & 2D & Appearance defects such as scratches, uneven color, and stains. \\
\hline
Multi-Spectral Camera & 2D & Surface material defects such as material aging and internal abnormalities such as poor welding. \\
\hline
Depth Camera & 3D & Geometric and assembly defects such as protrusions and damages. \\
\hline
LiDAR & 3D & Large-scale deformation defects such as structural offset and positional abnormalities. \\
\toprule[1.5pt]
\end{tabular}
}
\label{sensors}
\end{table}

\subsection{Defect Detection System}
\begin{figure*}[h!]
\centering\includegraphics[width=\linewidth]{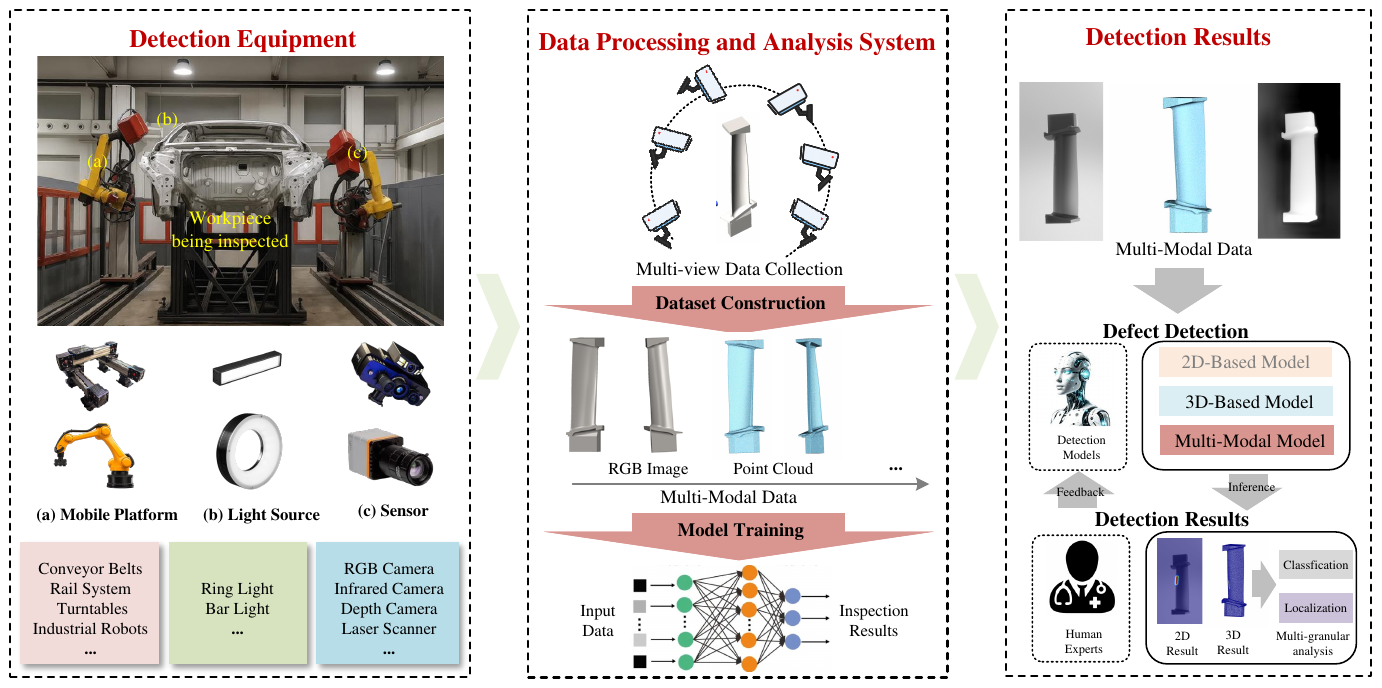}
\caption{Overview of a typical industrial defect detection system. The system consists of three main components: (1) Detection Equipment, including mobile platforms (e.g., conveyor belts, robotic arms), light sources (e.g., ring and bar lights), and various sensors (e.g., RGB cameras, depth sensors, laser scanners); (2) Data Processing and Analysis System, where multi-view data are collected to construct multi-modal datasets (e.g., RGB images, point clouds) for model training; and (3) Detection Results, where trained 2D-, 3D-, or multi-modal models perform defect detection and output classification or localization results.}
\vspace{-3mm}
\label{fig:system}
\end{figure*}

The detection system, depicted in Fig.~\ref{fig:system} (a), integrates several essential elements. Initially, imaging devices are chosen based on the specific attributes of the inspection target and the nature of anticipated defects, as outlined in Table~\ref{sensors}. Simultaneously, illumination conditions are adjusted to align with material properties, optimizing image contrast to improve defect visibility. To ensure exhaustive coverage and ideal sensor perspectives, an appropriate kinematic configuration is adopted, frequently incorporating combinatorial motion mechanisms to fulfill the diverse demands of industrial contexts, as detailed in Table~\ref{platform}. Subsequently, the data processing and analysis system orchestrates path and viewpoint planning, harmonizing the imaging devices and motion system to achieve thorough spatial sampling of the target. The acquired data are meticulously compiled into a structured database, encompassing reference templates of defect-free samples and annotated defective instances for algorithmic training. Finally, specialized inspection algorithms are implemented to identify anomalies and pinpoint defects, with the resulting insights serving to enhance product quality or streamline production processes.

\subsection{Defect Detection Datasets}
Public datasets have emerged as vital tools for validating and refining algorithmic performance, significantly expediting the evolution of defect detection models. Table~\ref{datasets} provides an overview of key datasets prevalent in industrial DD. In recent years, the research focus has transitioned from conventional closed-set DD~\citep{DAGM2007} to more intricate open-set scenarios~\citep{MVTec-AD}, responding to the pressing need to detect previously unseen defects in real-world industrial settings.

Concurrently, advancements in data modalities~\citep{MVTec3D,Real-IAD3} have propelled further refinements in detection techniques. Early investigations predominantly emphasized image-based anomaly detection~\citep{BTAD,MVTec-AD}, while the subsequent integration of 3D point cloud data has enriched geometric detail~\citep{MVTec3D,real3d,shape_anomaly}, enabling more accurate defect localization and characterization. More recently, the adoption of multimodal data---including RGB images, point clouds, and infrared imagery~\citep{MulSenAD})---has gained prominence. By capitalizing on the complementary capabilities of diverse sensors, these multimodal approaches yield a more complete representation of the target, thereby enhancing the robustness and adaptability of detection models amid the complexities of industrial environments.

Furthermore, the design of testing datasets has evolved to more accurately reflect the multifaceted challenges of real-world applications. For instance, Real-IAD~\citep{real-iad} and RAD~\citep{RAD} explicitly incorporate viewpoint variations, while Eyecandies~\citep{eyecandies} accounts for differing lighting conditions. Additionally, $M^{2}$AD~\citep{M2AD} simultaneously considers the effects of viewpoint and illumination. Such developments play a crucial role in aligning academic research with practical industrial needs.

\subsection{Defect Detection Challenges}
In real-world industrial DD applications, several obstacles impede the attainment of accurate detection outcomes, as elaborated below.

\noindent\textbf{Clear Imaging of Complex Industrial Objects:} Industrial components often feature intricate geometries and a variety of surface materials, such as metallic, transparent, or highly reflective finishes. This heterogeneity poses significant obstacles for imaging systems, which must resolve fine structural details under inconsistent lighting and viewpoint conditions.

\noindent\textbf{Visibility of Defects Across Modalities:} 2D modalities, such as RGB and infrared images, deliver high-resolution texture and appearance information, proving effective for surface anomaly detection. However, they lack depth perception and are vulnerable to variations in lighting and viewpoint. In contrast, 3D modalities, including LiDAR, structured light, and stereo vision, capture geometric and topological details, facilitating the identification of structural defects invisible in 2D. Yet, these 3D approaches are hampered by sparsity, occlusion, and noise.

\noindent\textbf{Subtle, Varied, and Novel Defect Types:} In industrial contexts, defects often manifest subtly---such as micro-scratches, minute dents, or faint deformations---while exhibiting considerable diversity and unpredictability, complicating detection efforts.

\noindent\textbf{Challenges in Data Annotation:} Annotating defect data is a resource-intensive process necessitating expert precision for accurate localization, incurring substantial costs. Moreover, the scarcity and imbalance of such data---stemming from the predominance of defect-free products---result in datasets skewed toward normal samples, with limited defective instances available to train robust supervised models.

\noindent\textbf{Demands for Precision and Speed:} Industrial inspection systems require exceptional accuracy alongside strict real-time performance to integrate seamlessly into production lines. Detection errors can precipitate significant downstream effects, such as quality deterioration or operational delays, while processing pipelines must maintain high frame rates---often exceeding 20--30 frames per second---to support high-throughput settings.

\noindent\textbf{Ensuring Robustness and Consistency:} Detection methods must withstand diverse sources of variability, including lighting fluctuations, sensor noise, part positioning, and manufacturing tolerances. Inconsistent outcomes across similar samples or over time can undermine system reliability and confidence in automated quality control, making consistency and generalizability across operational conditions imperative.

\begin{table*}[t]
\caption{
Summary of representative public benchmarks for industrial anomaly detection.
All datasets are categorized by year, modality, and task type. "PC" refers to point clouds. "Real" indicates whether any part of the dataset is real-world collected.
}
\vspace{2mm}
\label{datasets}
\scriptsize
\centering
\renewcommand\arraystretch{1.3}

% \begin{tabular}{lccccccccc}
% \begin{tabularx}{\textwidth}{@{\extracolsep{\fill}}lccccccccc}
\resizebox{\linewidth}{!}{
\begin{tabular}{lccccccccc }

\toprule
\textbf{Dataset} & \textbf{Year} & \textbf{Venue} & \textbf{Modality} & \textbf{\#Classes} & \textbf{\#Instances} & \textbf{\#Normal} & \textbf{\#Abnormal} & \textbf{Task} & \textbf{Real} \\
\midrule
DAGM 2007~\cite{DAGM2007}               & 2007 & DAGM Sym. & Image             & 10 & 11,500  & 10,000  & 1,500   & Close-Set & \ding{55} \\
MVTec~\cite{MVTec-AD}                   & 2021 & IJCV      & Image             & 15 & 5,354   & 4,096   & 1,258   & Open-Set  & \ding{51}  \\
BTAD~\cite{BTAD}                        & 2021 & ISIE      & Image             & 3  & 2,830   & 2,540   & 290     & Open-Set  & \ding{51}  \\
MPDD~\cite{MPDD}                        & 2021 & ICUMT     & Image             & 6  & 1,346   & 1,064   & 282     & Open-Set  & \ding{51}  \\
VisA~\cite{VisA}                        & 2022 & ECCV      & Image             & 12 & 10,821  & 9,621   & 1,200   & Open-Set  & \ding{51}  \\
MVTec LOCO~\cite{MVTecLOCO}             & 2022 & IJCV      & Image             & 5  & 3,340   & 2,347   & 993     & Open-Set  & \ding{51}  \\
PAD~\cite{PAD}                          & 2023 & NeurIPS   & Image             & 20 & 10,133  & 5,231   & 4,902   & Open-Set  & \ding{51}  \\
MSC-AD~\cite{MSC-AD}                    & 2024 & TII       & Image             & 12 & 9,720   & 8,640   & 1,080   & Open-Set  & \ding{51}  \\
Real-IAD~\cite{real-iad}                & 2024 & CVPR      & Image             & 30 & 151,050 & 99,721  & 51,329  & Open-Set  & \ding{51}  \\
VISION~\cite{vision}                    & 2024 & Arxiv     & Image             & 14 & 18,422  & 4,618   & 13,804  & Close-Set & \ding{51}  \\
WFDD~\cite{WFDD}                        & 2024 & ECCV      & Image             & 4  & 4,101   & 3,860   & 241     & Open-Set  & \ding{55} \\
VAD~\cite{VAD}                          & 2024 & CVPR      & Image             & 1  & 5,000   & 3,000   & 2,000   & Close-Set & \ding{51}  \\
RAD~\cite{RAD}                          & 2024 & IJCV      & Image             & 13 & 4,287   & 1,224   & 3,063   & Open-Set  & \ding{51}  \\
3CAD~\cite{3CAD}                        & 2025 & AAAI      & Image             & 8  & 27,039  & 15,577  & 11,462  & Open-Set  & \ding{51}  \\
MVTec2~\cite{MVTec2}                    & 2025 & Arxiv     & Image             & 8  & 8,004   & 4,705   & 3,299   & Open-Set  & \ding{51}  \\
$M^{2}$AD~\cite{M2AD}                   & 2025 & Arxiv     & Image             & 10 & 119,880 & 69,070  & 50,810  & Open-Set  & \ding{51}  \\
MANTA~\cite{MANTA}                      & 2025 & CVPR      & Image \& Text     & 38 & 686,690 & 652,455 & 34,235  & Open-Set  & \ding{51}  \\
Real3D-AD~\cite{real3d}                 & 2023 & NeurIPS   & PC                & 12 & 1,254   & 652     & 602     & Open-Set  & \ding{51}  \\
Anomaly-ShapeNet~\cite{shape_anomaly}  & 2024 & CVPR      & PC                & 40 & 1,600   & 160     & 1,440   & Open-Set  & \ding{55} \\
MiniShift~\cite{Simple3D}               & 2025 & Arxiv     & PC                & 12 & 2577    & 1137    & 1200    & Open-Set  &  \ding{55} \\
MVTec 3D~\cite{MVTec3D}                 & 2021 & VISAPP    & Image \& PC        & 10 & 3,852   & 2,904   & 948     & Open-Set  & \ding{51}  \\
Eyecandies~\cite{eyecandies}           & 2022 & ACCV      & Image \& PC       & 10 & 15,500  & 13,250  & 2,250   & Open-Set  & \ding{55} \\
Real-IAD D$^{3}$~\cite{Real-IAD3}       & 2025 & CVPR      & Image \& PC       & 20 & 8,450   & 5,000   & 3,450   & Open-Set  & \ding{51}  \\
MulSenAD~\cite{MulSenAD}               & 2025 & CVPR      & Image \& PC \& IR & 15 & 2,035   & 1,541   & 494     & Open-Set  & \ding{51}  \\
\bottomrule
\end{tabular}

}
\end{table*}

\section{2D Defect Detection}
\label{sec:image_defect_detection}

\begin{figure*}[ht!]
    \centering
    \includegraphics[width=\linewidth]{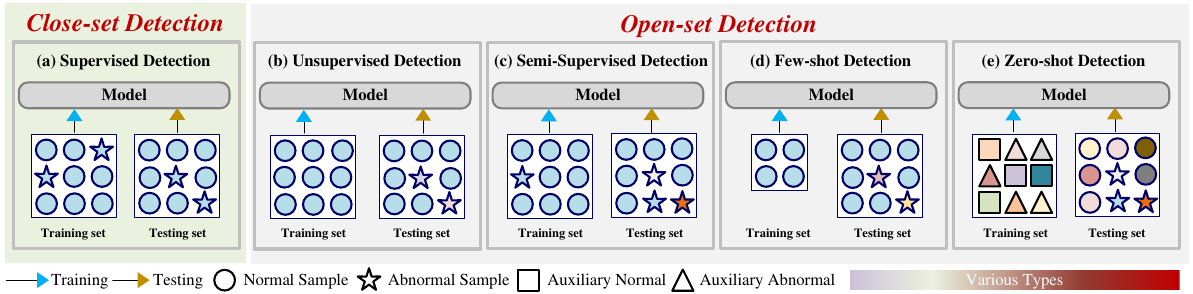}
    \caption{\textbf{Schematic of close-set and open-set DD paradigms. }(a) Close-set detection, assuming all defect types are known at training. (b) Unsupervised open-set detection, trained solely on normal samples. (c) Semi-supervised open-set detection, leveraging a limited number of anomalous samples with abundant normal samples during training. (d) Few-shot open-set detection, accessing only a small set of normal samples (typically $<8$) for training. (e) Zero-shot open-set detection, utilizing auxiliary normal and abnormal data for training and evaluating on novel classes.}
    \vspace{-3mm}
    \label{fig:2d_dd_scheme}
\end{figure*}

\begin{figure*}[h!]
\centering\includegraphics[width=\linewidth]{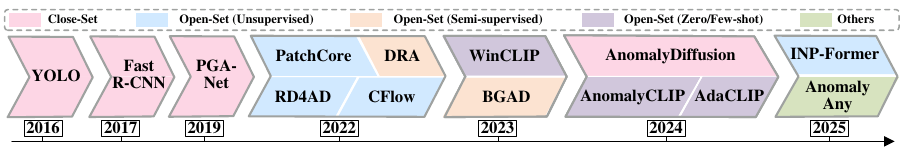}
\caption{\textbf{Timeline of representative methodologies in 2D DD.} In recent years, there has been a growing interest in open-set DD, accompanied by the emergence of various subtasks such as semi-supervised DD and zero-/few-shot DD.}
\vspace{-3mm}
\label{fig:2DDD}
\end{figure*}

2D DD constitutes a pivotal element of contemporary visual inspection systems, tasked with identifying and classifying imperfections within digital imagery~\cite{ameri2024systematic}. As illustrated in Fig.~\ref{fig:2d_dd_scheme} (a), initial investigations in this field predominantly focused on scenarios where training data encompassed all conceivable defect categories, facilitating the development of supervised models adept at recognizing and categorizing known anomalies. This approach, termed \textit{close-set DD}, assumes comprehensive prior knowledge of defect types. However, industrial settings frequently present complexities that challenge this paradigm, including the scarcity of defect samples~\cite{cao2024survey} and the emergence of previously unseen defect categories absent from training datasets. Such challenges have catalyzed the advent of \textit{open-set DD}, depicted in Fig.~\ref{fig:2d_dd_scheme} (b)$\sim$(e), which addresses the identification of novel defects not represented during model training. The representative 2D DD methods are shown in Fig.~\ref{fig:2DDD}.

\subsection{Close-Set 2D Defect Detection}
\label{subsec:close_set}

Close-set DD has evolved in tandem with advancements in computer vision, leveraging foundational developments in deep learning architectures. Early methodologies employed CNNs, with seminal works such as AlexNet~\cite{Krizhevsky_2017} marking a transition from traditional feature engineering to data-driven representation learning. Subsequent innovations, including Fast R-CNN~\cite{Fast_R_CNN} and Faster R-CNN~\cite{Faster_R_CNN}, enhanced detection accuracy by integrating region proposal networks, while single-stage detectors like the YOLO series~\cite{Redmon_2016,wang2024yolov10} prioritized real-time performance. More recently, the introduction of Vision Transformers (ViTs)~\cite{ViT} and ViT-based foundation models~\cite{oquab2023dinov2} has further elevated detection capabilities through attention-driven feature extraction.

Close-set DD systems operate across multiple levels of granularity, encompassing classification, localization via bounding boxes, and pixel-level segmentation. Initial approaches adapted object detection frameworks to industrial contexts~\cite{Gao_2022}, addressing challenges such as precision and robustness. The progression of computer vision technologies has significantly bolstered close-set DD, enabling systems to satisfy stringent industrial requirements. Nevertheless, several persistent challenges impede optimal performance: 1) \textbf{Subtle Defects}: Anomalies exhibiting low contrast with background regions demand high-sensitivity feature extraction to differentiate defects from noise effectively. 2)  \textbf{Multi-Scale Defects}: Variations in defect size and spatial distribution necessitate models capable of aggregating features across multiple scales. 3) \textbf{Real-Time Constraints}: High-throughput industrial environments impose stringent requirements on computational efficiency and inference speed. 4) \textbf{Data Scarcity}: Limited availability of annotated defect samples, particularly for rare categories, restricts model generalization and robustness.

\noindent\textbf{Addressing Subtle Defects.}
Detecting subtle defects requires advanced feature extraction techniques, with attention mechanisms playing a central role~\cite{Liu_2024,10887538}. Channel attention modules~\cite{Tang_2024} enhance global context by modeling inter-channel dependencies, whereas spatial attention mechanisms~\cite{Yuan_2024} prioritize salient regions, directing focus toward potential anomalies. Specialized models, such as BAF-Detector~\cite{Su_2022}, integrate bidirectional attention with feature pyramid networks to capture long-range dependencies and improve multi-scale detection. Similarly, DWWA-Net~\cite{Sui_2025} employs multiview attention to emphasize defect-relevant regions. Structural information further aids detection, with methods like those proposed by Shao et al.~\cite{Shao_2022} utilizing mean teacher frameworks to exploit structural variations, and Huang et al.~\cite{Huang_2024} leveraging skeleton-based representations.

\noindent\textbf{Addressing Multi-Scale Defects.}
To accommodate multi-scale defects, feature pyramid networks remain foundational, facilitating hierarchical feature aggregation~\cite{He_2020,Zhang_2024}. Recent advancements, such as MSC-DNet~\cite{Liu_2023}, incorporate dilated convolutions to expand receptive fields efficiently, while spatial pyramid pooling~\cite{Zhu_2023} captures multi-scale contextual information. Deformable convolutions~\cite{Peng_2024} dynamically adjust receptive fields to accommodate irregular defect morphologies, enhancing adaptability. These developments underscore the critical role of multi-scale feature extraction in complex industrial scenarios.

\noindent\textbf{Addressing Real-Time Constraints.}
Real-time performance is achieved through computationally efficient techniques, such as depthwise separable convolutions~\cite{Wang_2024,Cao_2024}, which reduce parameter complexity while preserving accuracy. Lightweight architectures like MobileNet~\cite{howard2017mobilenets} provide viable solutions for resource-constrained settings~\cite{Zhang_2024}. Knowledge distillation~\cite{Zhou_2024} further optimizes efficiency by transferring knowledge from large, high-capacity models to compact counterparts, maintaining performance with reduced latency.

\noindent\textbf{Addressing Data Scarcity.}
Data scarcity poses a significant barrier, particularly when annotated samples are limited or imbalanced. PGA-Net~\cite{Dong_2020} mitigates this by fine-tuning pre-trained networks for DD, achieving robust performance with minimal data. Bo et al.~\cite{Bo_i__2021} employ coarse-to-fine supervision to leverage weakly labeled samples, while CS-ResNet~\cite{Zhang_2021} uses weighted loss functions to address class imbalance. Contrastive learning approaches~\cite{Wan_2022,Wan_2024} enhance feature discrimination between normal and anomalous patterns. Domain adaptation techniques, such as one-shot style transfer~\cite{Ma_2023}, augment training data by transferring knowledge across domains, and incremental learning methods~\cite{Sun_2024,Zhao_2024} enable adaptation to new data without catastrophic forgetting.

\noindent\textbf{Defect Generation for Data Augmentation.}
Defect generation has emerged as a vital strategy to augment limited datasets. Early efforts utilized generative adversarial networks (GANs)~\cite{Liu_2020} to synthesize defect patterns. Recent few-shot learning approaches, such as DFMGAN~\cite{Duan_2023}, combine defect-free backbones with defect-aware residuals. Diffusion models, including AnomalyDiffusion~\cite{Hu_2024} and Defect-Gen~\cite{Yang_2024}, refine pre-trained models to generate precise defects, while frameworks like AnoGen~\cite{Gui_2024} and DualAnoDiff~\cite{jin2024dualanodiff} produce realistic anomalies from minimal samples. DefectFill~\cite{song2025defectfill} enhances inpainting with tailored loss functions, ensuring contextual coherence. For a detailed review of these generation methods, refer to Xu et al.~\cite{xu2025survey}.

\noindent\textbf{Remark.}
Close-set DD has achieved widespread adoption in industrial applications, benefiting from synergies with mainstream computer vision research. However, the shift toward small-batch, multi-variety production complicates the acquisition of large-scale labeled datasets, particularly during early production stages. This discrepancy exposes a critical limitation: close-set methods struggle to address \textit{unseen defect types}, which frequently arise in dynamic industrial contexts, necessitating advancements in open-set methodologies.

\subsection{Open-set 2D Defect Detection}

To address the challenges posed by the unpredictable nature of defect types in industrial inspection, numerous open-set DD methodologies have been proposed. Early approaches primarily relied on template matching~\cite{yoon_effective_2008}, a technique that leverages a reference template derived from normal product samples. Defects are identified by comparing this template with test images, mimicking human visual inspection processes. While subsequent efforts~\cite{zhang_soft_2021} have sought to enhance matching accuracy, template-based methods remain constrained by inherent robustness limitations, particularly due to intra-class variations in normal patterns and discrepancies in imaging conditions between template and test images. 

In contrast, anomaly detection methodologies~\cite{cao2024survey}, which aim to learn the distribution of normal data from extensive datasets (typically comprising only normal samples), have demonstrated superior robustness to distribution shifts. This paradigm has garnered considerable attention in recent years, particularly with the availability of specialized benchmark datasets such as MVTec AD~\cite{MVTec-AD}, VisA~\cite{VisA}, and Real-IAD~\cite{real-iad}. While unsupervised anomaly detection~\cite{Uninformed_Students}---which learns exclusively from normal data---remains the dominant framework, recent research has increasingly explored alternative approaches, including semi-supervised anomaly detection~\cite{BGAD,BiaS}, zero/few-shot anomaly detection~\cite{WinCLIP,AdaCLIP,AnomalyCLIP}, etc. Below, we provide a structured overview of these methodologies.

\subsubsection{Unsupervised Anomaly Detection}
Unsupervised anomaly detection, as shown in Fig.~\ref{fig:2d_dd_scheme} (b), operates under the assumption that abundant normal samples are typically available during the training phase, while abnormal samples (e.g., defective instances) are often scarce or difficult to collect. Consequently, training is conducted exclusively on normal samples. Methodologies in this domain can generally be structured into two distinct stages: \textit{feature extraction}, which focuses on deriving discriminative representations from input data, and \textit{feature comparison}, which evaluates the deviation of test samples from learned normal distributions.

The \textit{feature extraction paradigm} has evolved significantly over time. Early methodologies predominantly relied on handcrafted features~\cite{xie_texems_2007} or self-supervised representations~\cite{CutPaste,NSA}, which were constrained by their dependency on domain-specific engineering and labeled data. Modern approaches, however, have increasingly turned to leveraging pre-trained features~\cite{He_2016}, particularly those derived from large-scale foundational models~\cite{oquab2023dinov2}, which demonstrate superior generalization and scalability across diverse applications. For instance, Dinomaly~\cite{Dinomaly} utilizes Dino~\cite{oquab2023dinov2} for feature extraction. Notwithstanding their advantages, even state-of-the-art (SOTA) pre-trained networks---foundational or otherwise---often exhibit inherent domain shifts when deployed in industrial inspection contexts. To mitigate this challenge, recent advancements have emphasized the fine-tuning of pre-trained architectures to better align with target-domain requirements.

Notably, {EfficientAD}~\cite{EfficientAD} distills features from ImageNet-pretrained networks into lightweight architectures with constrained receptive fields, thereby optimizing computational efficiency while preserving detection accuracy. In parallel, {CFA}~\cite{CFA} introduces a learnable patch descriptor that employs contrastive learning to embed target-oriented semantic features, while {REB}~\cite{REB} synthesizes defective image proxies via self-supervised tasks to facilitate domain-adaptive classifier training. Concurrently, {ReContrast}~\cite{ReContrast} proposes end-to-end network optimization to minimize pre-training domain bias and recalibrate model focus toward target scenarios. This strategy is echoed in {UniNet}~\cite{UniNet}, which employs contrastive learning frameworks to enhance feature distinguishability, thereby addressing domain misalignment while maintaining computational efficiency. Collectively, these methodologies underscore the growing emphasis on domain-adaptive feature refinement as a critical pathway for advancing industrial inspection systems.

The \textit{feature comparison stage}, on the other hand, encompasses a diverse set of methodologies, including regression-based~\cite{IKD1,RD4AD}, memory bank-based~\cite{PatchCore,GCPF}, flow-based~\cite{cflow,MSFlow,yao_localglobal_2024}, and discrimination-based~\cite{SimpleNet,GLASS} approaches. Each category addresses the challenge of quantifying deviations from normality through distinct algorithmic frameworks. The subsequent sections provide a comprehensive overview of these methodologies.

\noindent \textbf{Regression-based Methods:} Regression-based anomaly detection frameworks leverage the principle of reconstructing or regressing normal features to identify deviations from expected patterns. These approaches typically involve training two interconnected networks: a pre-trained, high-capacity network that generates discriminative feature representations and a secondary network trained to regress these features during the optimization phase. Since the regression network is exclusively exposed to normal samples during training, it tends to produce larger reconstruction errors when processing anomalous inputs, thereby making the regression error a critical indicator for anomaly detection. This family of methods is broadly categorized as knowledge distillation-based or reconstruction-based techniques, depending on their specific architectural design.

Over the years, significant advancements have been made to enhance the effectiveness of regression-based anomaly detection. Early approaches, such as knowledge distillation with parallel networks~\cite{STPFM,MRKD}, utilized dual networks to capture and regress feature representations. Subsequent improvements introduced multiple parallel network pairs~\cite{Uninformed_Students} or incorporated trainable projection heads alongside pre-trained networks~\cite{PFM,PEFM}. Sequential architectures, such as multi-scale feature reconstruction via CNNs~\cite{DFR} or reverse ResNet designs in RD4AD~\cite{RD4AD}, further refined the reconstruction process by leveraging hierarchical feature hierarchies. Hybrid architectures, such as those combining flow-based and CNN-based networks~\cite{AST}, expanded the scope of feature representation, enabling more robust anomaly detection.

More recently, transformer-based architectures have gained prominence in this domain. Methods like UTRAD~\cite{UTRAD} and UniAD~\cite{UniAD} leverage vision transformers to reconstruct multi-scale features in a sequential manner, capitalizing on the transformer's inherent ability to capture long-range dependencies and contextual information. Diffusion models have also been integrated into this framework, with techniques such as pixel-level and feature-level reconstruction errors~\cite{DiAD} or multi-scale strategies~\cite{MTDiff} to enhance anomaly detection performance. Notably, DeCo-Diff~\cite{beizaee_correcting_2025} introduces a novel perspective by modeling anomalies as latent noise, focusing transformations exclusively on anomalous regions while preserving normal feature spaces.

State space models, such as MambaAD~\cite{Mambaad}, offer an alternative approach by reconstructing features through a sequential architecture similar to RD4AD, emphasizing temporal coherence in feature representation. Additionally, some methods generate multiple alternative inputs to enrich the feature space. For instance, Yan et al.~\cite{yan_unsupervised_2021} generate images at different granularity levels before applying regression, while others reconstruct images from a multi-frequency perspective~\cite{liang_omni-frequency_2023,AFSC} to capture diverse feature representations. These advancements collectively highlight the evolving landscape of regression-based anomaly detection, driven by innovations in network design, feature reconstruction strategies, and the integration of advanced architectures such as transformers and diffusion models.

Despite architectural variations, a pervasive challenge in anomaly detection frameworks is the phenomenon of \textit{overgeneralization}~\cite{CDO}, wherein trained models inadvertently generalize to abnormal data manifolds. This results in diminished regression errors for anomalous inputs, thereby undermining the efficacy of detection mechanisms. To mitigate this issue, several methodologies have been proposed, with a common strategy involving the incorporation of normal pattern representations to guide regression processes, often embedded within reconstruction-based frameworks.

Among the pioneering approaches, {MemAE}~\cite{MemAE} addresses this challenge by storing prototypical normal patterns and substituting extracted latent vectors with the nearest normal feature representations. Expanding upon this concept, {DMAD}~\cite{liu_diversity-measurable_2023} introduces a {Pyramid Deformation Module} to model diverse normal patterns and quantify anomaly severity through multi-scale deformation field estimation between reconstructed references and original inputs. Similarly, {PMB}~\cite{xing_visual_2023} proposes a {Partition Memory Bank} module that learns and stores granular features of normal samples while preserving semantic coherence through a novel partitioning mechanism and query generation strategy. This approach enhances the memory module's representational capacity by maintaining contextual information.
{THFR}~\cite{guo_template-guided_2023} and TFA-Net~\cite{luo2024template} employ a template-guided compensation strategy, restoring distorted features by identifying the most similar normal images to guide reconstruction. {HVQ-Trans}~\cite{lu_hierarchical_2023} extends the MemAE framework by quantizing and storing normal feature representations through hierarchical processing. Meanwhile, {MemKD}~\cite{gu_remembering_2023} adopts a parallel architecture leveraging knowledge distillation to improve the generalization of normal pattern representations.
{NDP-Net}~\cite{luo_normal_2023} explicitly incorporates normal images as reference inputs and introduces a reference-based attention mechanism to guide reconstruction. Focusing on anomaly localization, {FOD}~\cite{FOD} innovates by transforming self-attention maps into a dual-branch architecture that explicitly models intra- and inter-image correlations to highlight abnormal patterns. More recently, {PNPT}~\cite{PNPT} and {RLR}~\cite{he_learning_2024} utilize prompting learning techniques to automatically derive normal prompts, while {INP-Former}~\cite{INP-Former} extracts {Intrinsic Normal Prototypes} (INPs) directly from input images to provide more aligned and concise representations of normality, achieving superior performance across a wide range of anomaly detection scenarios and tasks.

Another prominent category of solutions addressing over-generalization in anomaly detection explicitly incorporates synthetic anomalies into the training pipeline, thereby reframing the reconstruction task as a restoration-based framework. This paradigm leverages the principle that training neural networks to reconstruct normal patterns from synthetically corrupted inputs can enhance discriminative feature learning between normal and abnormal data distributions.  For instance, Dual-Siamese \cite{tao_unsupervised_2022} decomposes the restoration process into two complementary sub-networks: one specialized in reconstructing normal patterns, and the other focused on restoring synthetically anomalous regions to their normal counterparts. Similarly, RD++ \cite{RD++} strengthens feature discriminability through explicit anomaly injection, while OmniAL \cite{zhao_omnial_2023} employs a cascaded approach combining normal pattern generation with subsequent anomaly localization refinement. DeSTSeg \cite{destseg} integrates knowledge distillation principles by combining a pre-trained teacher network with a denoising student encoder-decoder and segmentation module, where synthetic anomaly masks guide precise localization. MRKD \cite{jiang_masked_2023} extends this framework by introducing hierarchical perturbations at both image and feature levels. Concurrent approaches such as CDO and Pull\&Push \cite{CDO,zhou_pull_2023} optimize dual objectives: minimizing reconstruction error for normal patterns while maximizing error for synthetic anomalies. GeneralAD \cite{GeneralAD} employs elementary operations like noise injection and feature shuffling to synthesize pseudo-anomalous samples, whereas ADPS \cite{ADPS} utilizes asymmetric knowledge distillation with noise-augmented image inputs. RealNet \cite{RealNet} advances this paradigm through diffusion-based generation of subtle anomalies, thereby refining restoration fidelity.

Complementary research trajectories focus on abnormal feature masking during reconstruction, enabling networks to concentrate exclusively on normal feature spaces. Methodologies such as those proposed by \cite{madan_self-supervised_2024} implement predictive convolutional attentive mechanisms to reconstruct masked regions using contextual priors. RIAD \cite{RIAD} applies stochastic masking followed by inpainting, while subsequent works \cite{huang_self-supervised_2023,jiang_masked_2023-1} advance masking strategies to enhance contextual feature understanding. MSTAD \cite{MSTAD} extends this principle to feature space through random masking of pre-trained feature maps, while MLDFR \cite{MLDFR} combines masking with hybrid CNN-ViT architectures. AMINet \cite{AMINet} introduces adaptive masking mechanisms that dynamically suppress anomalous regions during reconstruction. One-for-All \cite{yao_one-for-all_2023} proposes prototype-guided proposal masking to systematically eliminate abnormal information, while PSA-VT \cite{yao_scalable_2024} employs token subset mutual prediction to enforce long-range contextual integration.

\noindent\textbf{Memory Bank-Based Methods:} 
This category of methods maintains a memory bank of features extracted from normal training data and compares test features against this stored memory. For example, {SPADE}~\cite{SPADE} stores all normal training features in a memory bank and employs K-nearest neighbor (KNN) to identify the closest normal features, leveraging distance metrics to compute anomaly scores. While this approach is intuitive and effective, it often encounters scalability challenges due to the prohibitive storage requirements associated with large-scale datasets.

To address these limitations, {GCPF}~\cite{GCPF} adopts a probabilistic framework by modeling normal features using multiple Gaussian distributions, thereby reducing computational overhead. Similarly, {PatchCore}~\cite{PatchCore} employs coreset sampling to select the most representative features from the training set, achieving significant efficiency improvements while preserving detection performance. Notably, PatchCore's elegant design has emerged as a landmark method, establishing a benchmark for subsequent approaches.

Building upon PatchCore's framework, recent advancements such as {ReConPatch}~\cite{ReConPatch}, {REB}~\cite{REB}, and {CFA}~\cite{CFA} have incorporated contrastive learning to enhance feature discriminability, further improving anomaly detection accuracy. {PaDiM}~\cite{PADIM} extends this approach by explicitly modeling position-specific features using Gaussian distributions, thereby preserving spatial information critical for detecting localized anomalies. Meanwhile, {PNI}~\cite{PNI} adopts a pixel-level strategy by constructing memory banks for individual positions, leveraging consistent patterns observed in normal images. {FastRecon}~\cite{FastRecon} diverges from traditional nearest-neighbor search by regressing testing features as linear combinations of memory bank features, offering a computationally efficient alternative.

\noindent \textbf{Flow-based Methods:} Flow-based methods utilize normalizing flows~\cite{rezende2015variational}, a class of generative models that transform complex probability distributions into simpler ones through invertible mappings. These approaches explicitly model the distribution of normal features, enabling anomaly detection by assigning low likelihoods to unseen anomalies. DifferNet~\cite{rudolph_same_2021} pioneered this framework by processing multiscale feature representations of input images through a feature extractor, concatenating the outputs, and modeling the resulting features via a normalizing flow. The model is trained using maximum likelihood estimation to capture the distribution of normal data.

Building on this foundation, CFlow-AD~\cite{cflow} introduces a multiscale feature pyramid constructed from feature maps at different levels, thereby improving computational efficiency and capturing diverse receptive fields. Additionally, CFlow-AD extends normalizing flows to pixel-wise anomaly localization by estimating the likelihood of each feature vector in the spatial domain. However, this approach processes feature vectors independently, neglecting spatial contextual information, which can result in fragmented localization outputs. To address this limitation, CSFlow~\cite{CSFlow} proposes a fully convolutional cross-scale flow module that jointly processes multiscale feature maps, preserving spatial context. Similarly, MSFlow~\cite{MSFlow} estimates features across all spatial positions in parallel, leveraging 3×3 convolutions to implicitly learn contextual relationships. CARF~\cite{liu_cross-attention_2024} further advances this paradigm by integrating an interactive cross-attention pattern flow module, which aligns the distributions of multi-layer features while retaining rich scale information. This is particularly advantageous for detecting small-scale defects that may be lost in high-level feature representations.

PyramidFlow~\cite{PyramidFlow} introduces volume normalization to preserve task-relevant implicit priors, enabling the training of backbone networks using contrastive learning rather than relying on pre-trained models, which can introduce domain-specific biases. THF~\cite{jiang_unsupervised_2024} proposes a masked flow mechanism, which enhances traditional convolutional kernels by masking partial regions to prevent excessive minimization of negative log-likelihood. Recognizing that input data may exhibit heterogeneous clustered structures, HGAD~\cite{yao_hierarchical_2024} maps inputs into distinct clusters using normalizing flows, thereby moving beyond the unified center assumption of prior methods. BGAD~\cite{BGAD} extends flow-based methods to semi-supervised anomaly detection by encouraging normal samples to exhibit higher likelihoods than abnormal ones, thereby leveraging limited labeled data for improved generalization.

Recent advancements have integrated regression-based techniques with normalizing flows. For instance,~\cite{yao_dual-attention_2024} utilizes normalizing flows to estimate anomaly likelihoods by quantifying discrepancies between pre-trained features and reconstruction results. 

\noindent \textbf{Discrimination-based Methods:}
These methods center on training a discriminative network to directly identify anomalous features through the use of synthetic pseudo-anomalies. Early approaches, such as CutPaste~\cite{CutPaste}, rely on stochastic patch manipulation within images to generate pseudo-anomalies. While this technique trains a discriminative network to localize synthetic irregularities, its reliance on overly simplistic pseudo-anomalies imposes significant limitations on capturing realistic irregularities. To address this limitation, NSA~\cite{NSA} integrates Poisson image editing to seamlessly integrate patches of varying sizes from different images, thereby generating more realistic and diverse synthetic anomalies that better approximate real-world sub-image irregularities.

Subsequent advancements transition from image-level perturbations to feature-level synthesis. SimpleNet~\cite{SimpleNet} introduces adversarial noise directly into the feature space, thereby enforcing compact decision boundaries for normal data. Building on this foundation, PBAS~\cite{chen_progressive_2024} refines the approach by directionally synthesizing critical feature-level anomalies while maintaining a tightly clustered normal distribution. GLASS~\cite{GLASS} further advances this paradigm by proposing a unified framework constrained by manifold and hypersphere distributions, enabling comprehensive anomaly coverage at both feature and image levels.

Recent innovations explore reconstruction-based strategies to enhance discriminative learning capabilities. DRAEM \cite{DRAEM} reconstructs inputs into their nominal counterparts before feeding both the original and reconstructed images into a discriminative network trained to distinguish discrepancies between the two. This approach is extended in subsequent works, such as~\cite{wang_produce_2024} and~\cite{xing_normal_2024}, which reconstruct features directly and incorporate reference features from normal distributions to improve segmentation performance. Notably,~\cite{zhang_unsupervised_2023} leverages diffusion models for high-fidelity reconstruction, demonstrating that advanced generative techniques can significantly enhance the discriminative network's capacity to detect anomalies. Taken together, these approaches demonstrate a progression from basic pseudo-anomaly generation techniques toward advanced feature synthesis and reconstruction methods, thereby emphasizing the essential contribution of multi-level anomaly representation within discrimination-based frameworks.

\subsubsection{Semi-supervised Anomaly Detection}

In practical industrial inspection systems, while the acquisition of sufficient abnormal samples frequently poses significant challenges due to their inherent rarity and stochastic occurrence patterns, limited labeled abnormal data are often available, motivating the adoption of semi-supervised anomaly detection frameworks, as shown in Fig.~\ref{fig:2d_dd_scheme} (c). These methodologies strategically integrate scarce labeled abnormal instances with abundant normal samples to enhance detection performance beyond the capabilities of purely unsupervised approaches. This paradigm, which leverages limited anomaly information to improve generalization to unseen defect patterns, is commonly termed open-set anomaly detection~\citep{DRA,DevNet}.

Prior anomaly detection techniques like DeepSVDD \citep{DeepSVDD} have demonstrated efficacy by refining decision boundaries through hypersphere optimization using abnormal samples. Unsupervised anomaly detection frameworks employing synthetic pseudo-anomalies can also be systematically extended to semi-supervised settings by substituting synthetic anomalies with authentic labeled instances.  Despite this progress, recent advancements have prioritized the development of tailored architectures and training protocols explicitly designed for semi-supervised scenarios.

For instance, DevNet~\citep{DevNet} introduces a novel framework that optimizes anomaly scores via neural deviation learning, wherein labeled anomalies and prior probability constraints enforce discriminative representations of normal data while ensuring statistically significant deviations for anomalous instances. Building on this foundation, SegAD~\citep{SegAD} leverages statistical features extracted from anomaly maps generated by unsupervised detectors to train a supervised classifier, thereby improving localization accuracy. SuperSimpleNet~\citep{SuperSimpleNet} extends SimpleNet by replacing synthetic pseudo-anomalies with real labeled anomalies, enhancing model robustness against seen defects. DRA~\citep{DRA} advances this approach by disentangling known and unknown anomalies through dual learning heads and generating unknown anomalies via pseudo-paste techniques. Conversely, BiaS~\citep{BiaS} extends regression-based methods such as CDO~\citep{CDO} by introducing dual networks to simultaneously address known and unknown anomalies. Instead, BGAD~\cite{BGAD} extends flow-based methods into this semi-supervised setup with a push-and-pull learning objective. 
PRN~\citep{PRN} focuses on feature residuals to reconstruct anomaly segmentation maps, while DPDL~\citep{DPDL} constructs a compact distribution space for normal samples using Gaussian prototypes and diffusion learning, thereby improving anomaly separation. AHL~\citep{AHL} proposes simulating diverse anomaly distributions to adaptively enhance abnormality modeling in existing frameworks. RobustPatch~\citep{RobustPatch} further improves discriminability through a self-cross scoring mechanism and feature AutoEncoder, which assigns weighted anomaly likelihood scores to extracted features.

Collectively, semi-supervised anomaly detection methodologies largely represent extensions of existing unsupervised frameworks, aiming to preserve generalization to unseen anomalies while enhancing detection efficacy for seen anomalies. Defect generation techniques, offer complementary benefits by augmenting the diversity of seen anomalies, thereby further improving the robustness of semi-supervised approaches. This integration underscores the potential for hybrid methodologies to address the dual challenges of limited labeled data and complex industrial defect patterns.

\subsubsection{Zero/Few-shot Anomaly Detection}
Recent advances in anomaly detection have shifted focus from conventional approaches reliant on large-scale datasets toward developing data-efficient frameworks capable of operating under severe data scarcity constraints. This methodological reorientation addresses scenarios ranging from few-shot settings (Fig.~\ref{fig:2d_dd_scheme} (d)), where only minimal normal instances (\textit{e.g.}, 1-8 samples) are available, to zero-shot (Fig.~\ref{fig:2d_dd_scheme} (e)) configurations entirely devoid of target-category exemplars. Such frameworks prioritize resource-constrained generalization while preserving anomaly detection robustness, thereby challenging traditional assumptions about the necessity of extensive training data. This paradigm shift not only expands the applicability of anomaly detection systems to domains with inherently limited data but also represents a critical step toward operationalizing anomaly detection as a foundational model. 

\noindent \textbf{Vanilla Few-Shot Anomaly Detection:}  
Early frameworks extended traditional unsupervised methods by integrating memory-augmented mechanisms. For instance, RegAD~\cite{RegAD} employed spatial transformation networks to align test samples with reference prototypes, alleviating performance degradation in data-limited settings. Subsequent work, such as CAReg~\cite{CAReg}, enhanced category-specific representation through aggregated intra-class feature statistics, thereby reducing dependency on large-scale corpora. GraphCore~\cite{GraphCore} further advanced geometric invariance via isometric feature embedding, demonstrating superior generalization across morphological variations.  Component-wise decomposition strategies, such as those employed in UniVAD~\cite{UniVAD}, systematically decompose anomaly localization into granular feature correspondences, thereby enabling fine-grained detection under extreme data scarcity. 

Contemporary frameworks have increasingly adopted regression-based methodologies for anomaly localization tasks. Regression-based approaches, exemplified by One-to-Normal~\cite{One-to-Normal}, leverage diffusion-based generative models to synthesize anomaly-free reconstructions, quantifying deviations through pixel-wise residual analysis. DFD~\cite{DFD} advances this paradigm by transforming spatial representations into multi-frequency encodings, thereby enhancing the detectability of subtle morphological aberrations that remain obscured in conventional feature spaces.

Residual features, defined as the discrepancy between test representations and reference normal features, encode rich discriminative information about anomalies. Several methodologies have capitalized on this principle. InCTRL~\cite{InCTRL} systematically formalizes residual learning through unbiased text-conditioned representations, harmonizing detection pipelines across normal and abnormal modalities. ResAD~\cite{ResAD} enforces compactness in normal feature distributions via hyperspherical embedding of residual manifolds, while PCSNet~\cite{PCSNet} refines this approach through nearest-neighbor residual computation and discriminative manifold processing. MetaUAS~\cite{MetaUAS} introduces soft alignment mechanisms to address spatial misregistration, facilitating one-shot detection in purely synthetic training regimes.

To address intrinsic data limitations stemming from sample scarcity, novel augmentation strategies integrate external semantic knowledge into model training. Lee et al.~\cite{lee_text-guided_2024} proposed a prompt-generation architecture that synthesizes domain-specific normal samples via text-conditioned diffusion, thereby expanding training corpora without requiring additional labeled data.

%%% zero-shot
\noindent \textbf{Vanilla Zero-shot Anomaly Detection:}
Early approaches to zero-shot anomaly detection originated as adaptations of memory-bank frameworks, constrained by the absence of external normality references. Aota et al.~\cite{aota_zero-shot_2023} established this paradigm by formulating anomaly detection through patch-wise self-comparisons within single-image textures, thereby circumventing reliance on external normal data. Building on this foundation, Ardelean et al.~\cite{ardelean_high-fidelity_2024} introduced a bijective mapping derived from 1-dimensional Wasserstein distance, which enhanced anomaly localization precision by quantifying distributional discrepancies between patch representations.

Notwithstanding their efficacy in leveraging local contextual cues, these methods exhibit limitations in modeling global structural coherence. To address this, GRNR~\cite{GRNR} proposed a self-reconstructive regression framework that synergistically integrates local neighbor support for reconstruction with global normality constraints. This dual mechanism ensures both visually coherent reconstructions and adherence to canonical normality patterns. 

In a paradigmatic departure from conventional single-image protocols, MuSc~\cite{MuSc} introduces a zero-batch mechanism that leverages multi-input processing. By establishing inter-sample outlier scoring through batch-level comparisons, this approach assigns anomaly scores inversely proportional to pattern frequency across the batch.

\noindent \textbf{VLMs-based Methods}
The rapid advancement of Vision-Language Models (VLMs) has spurred significant interest in leveraging their zero-shot capabilities and feature extraction capabilities for anomaly detection. In this paradigm, few-shot and zero-shot anomaly detection share several conceptual similarities, particularly in their reliance on semantic alignment between visual and textual modalities. For instance, methods~\cite{WinCLIP} such as those incorporating VLMs-based memory banks have demonstrated efficacy in extending zero-shot frameworks to few-shot scenarios. 

WinCLIP~\cite{WinCLIP} pioneers zero-shot anomaly detection by leveraging manually crafted normal/abnormal text prompts, computing semantic similarities between these prompts and test images using CLIP. Tamura~\cite{tamura_random_2023} diverges from prompt-based strategies by training a feed-forward network to distinguish randomly generated normal and abnormal text embeddings extracted by CLIP, thereby circumventing direct prompt similarity calculations. MAYDAY~\cite{MAEDAY} exemplifies this approach by directly utilizing pre-trained Masked Auto Encoders (MAEs) to reconstruct unknown images to their normal reference patterns, quantifying reconstruction errors as anomaly scores. AnomalyVLM~\cite{AnomalyVLM} (also known as SAA) introduces a hybrid approach combining GroundingDINO~\cite{GroundingDINO} and SAM~\cite{SAM}, employing hybrid prompts to mitigate false alarm rates in zero-shot settings.
 
However, since existing VLMs are not inherently designed for anomaly detection, several methods focus on adapting CLIP to enhance its suitability for this task. APRIL-GAN~\cite{APRIL-GAN} (also referred to as VAND) introduces a learnable projection layer appended to pre-trained CLIP, trained using auxiliary annotated anomaly detection data and tested on unseen categories. CLIP AD~\cite{ClipAD} addresses challenges such as opposite predictions and irrelevant highlights by integrating multi-level features and applying the architectural and feature surgery strategy from CLIP Surgery~\cite{CLIP_Surgery}. AnomalyGPT~\cite{AnomalyGPT} extends CLIP's applicability to few-shot scenarios by aligning normal and synthetic anomalies to normal/abnormal text embeddings, while MVFA~\cite{MVFA} further refines this approach by inserting trainable visual adaptation layers within CLIP rather than appending adaptation layers. CMAD~\cite{mao2025beyond} further proposes to learn transferable visual prototypes and can even detect anomalies in unseen modalities. 

Prompt learning has emerged as a critical research direction, driven by the limitations of manually crafted prompts. AnomalyCLIP~\cite{AnomalyCLIP} proposes learning object-agnostic text prompts, while LTAD~\cite{LTAD} addresses the ambiguity of original class names by introducing learned pseudo class names. VCP-CLIP~\cite{VCP-CLIP} dynamically refines text prompts by incorporating input-specific information, and AdaCLIP~\cite{AdaCLIP} extends this idea by dynamically prompting both the image and text encoders. Bayes-PFL~\cite{Bayes-PFL} introduces a prompt flow module to learn both image-specific and image-agnostic distributions. One-for-All~\cite{One-for-All} and KAG-Prompt~\cite{KAG-prompt} generate dynamic prompts based on input reasoning, with the latter focusing on cross-layer relations among visual features. PromptAD~\cite{PromptAD} significantly improves few-shot performance by explicitly controlling the margin between normal and anomaly prompt features. Similarly to PromptAD, AA-CLIP~\cite{AA-CLIP} addresses CLIP's limited distinguishability between normality and abnormality by enhancing its anomaly-awareness capabilities. Collectively, these advancements underscore the growing sophistication of VLMs-based approaches in anomaly detection, while also highlighting the ongoing challenges in optimizing model adaptability and generalizability.

\noindent \textbf{MLLMs-based Methods:}
Recent advancements in Multimodal Large Language Models (MLLMs) have showcased significant zero-shot reasoning capabilities, prompting researchers to investigate their potential applications in anomaly detection. For instance, FADE~\cite{FADE} employs MLLMs to generate more informative prompts, thereby enhancing the interpretability and accuracy of anomaly detection systems. Building on this approach, LogSAD~\cite{LogSAD} integrates component-wise features and leverages GPT-4 for semantic-level matching, thereby transcending the limitations of traditional visual feature-based methods. LogiCode~\cite{LogiCode} further innovates by defining visual parse APIs and utilizing GPT-4 to automate coding with these APIs, enabling reasoning based on predefined product standards. FiLo~\cite{FiLo} advances the field by introducing fine-grained anomaly descriptions for each category, employing MLLMs to enhance detection accuracy and interpretability through adaptively learned textual templates. 

Recent advancements in the field have investigated the direct application of MLLMs in anomaly detection and reasoning. While initial investigations demonstrate that pre-trained MLLMs face challenges in anomaly comprehension~\cite{ALFA,MMAD}, a multi-modal prompting strategy has been proposed to integrate domain-specific expert knowledge and guide MLLM reasoning capabilities~\cite{xu2024custimizing}. Furthermore, Xu et al.~\cite{Anomaly-OV} introduced Anomaly-Instruct-125k, the first large-scale visual instruction tuning dataset designed to address anomaly reasoning challenges. Furthermore, they developed Anomaly-OV, a specialized visual assistant capable of performing zero-shot anomaly detection and reasoning. This work not only mitigates the inherent limitations of pre-trained MLLMs but also establishes a foundational framework for anomaly reasoning and downstream operations, such as anomaly localization and resolution. By addressing critical gaps in existing methodologies, this research contributes to advancing the theoretical and practical dimensions of anomaly detection systems, thereby establishing a transformative framework for advancing anomaly reasoning methodologies.

\subsubsection{Remark}

Practical industrial inspection scenarios present unique challenges that extend beyond the scope of conventional image anomaly detection frameworks, thereby necessitating specialized research avenues. A prominent issue stems from the sequential nature of detection tasks in production environments, where models are required to incrementally adapt to novel anomaly categories without the luxury of retraining on the entirety of historical data. The integration of continual learning into anomaly detection was pioneered by DNE~\cite{DNE}, which mitigates catastrophic forgetting during sequential training phases. Nonetheless, DNE's scope is confined to anomaly detection, omitting localization capabilities. This limitation was subsequently addressed by UCAD~\cite{UCAD}, which extends the framework to include anomaly localization while simultaneously reducing computational demands and further alleviating catastrophic forgetting. Building upon these foundations, One-for-More~\cite{one-for-More} introduces a diffusion-based continual learning paradigm that employs gradient projection techniques to enhance model stability across sequential anomaly detection tasks.

Another pressing challenge is noisy anomaly detection, wherein anomalous samples may inadvertently contaminate the normal training data. To bolster robustness in such scenarios, SoftPatch~\cite{SoftPatch} proposes generating patch-level outlier scores prior to coreset construction, thereby enhancing the model's resilience to noisy inputs. Extending this line of research, TailedCore~\cite{TailedCore} incorporates class-size prediction mechanisms, which further refine the anomaly detection process under noisy conditions.

A fundamental constraint in DD systems arises from the scarcity of training data for unseen anomaly categories. To circumvent this limitation, AnomalyAny~\cite{AnomalyAny} manipulates the attention matrix of Stable Diffusion, thereby enabling the generation of arbitrary anomalies on previously unseen normal images---a notable first in the field. In a divergent yet complementary approach, AnomalyPainter~\cite{AnomalyPainter} introduces auxiliary patterns as synthetic anomaly sources and employs ControlNet to integrate these patterns seamlessly into test images. Distinct from the defect generation methodologies discussed in Section~\ref{subsec:close_set}, which primarily aim to augment known anomaly types, both AnomalyAny and AnomalyPainter are explicitly designed to generate realistic and diverse unseen anomalies across novel product categories.

\subsection{Discussion}

The domain of image DD for intricate industrial components has witnessed substantial evolution, marked by a paradigm shift from traditional close-set approaches to more sophisticated open-set methodologies. Open-set DD has garnered increasing attention, particularly through the lens of unsupervised anomaly detection, which exclusively utilizes normal samples to characterize normative patterns. Various techniques, including regression-based, memory bank-based, flow-based, and discrimination-based methods, have each contributed distinct advantages, with contemporary innovations emphasizing domain-adaptive feature refinement to counteract overgeneralization tendencies. Furthermore, semi-supervised and zero/few-shot anomaly detection frameworks augment these capabilities by leveraging limited labeled data or functioning effectively under severe data paucity, respectively. The synergistic integration of advancements from both close-set and open-set DD paradigms, coupled with concerted efforts to address enduring challenges such as data scarcity and computational efficiency, holds the potential to significantly enhance the reliability and adaptability of visual inspection systems within complex manufacturing ecosystems.

\begin{figure*}[h!]
\centering\includegraphics[width=\linewidth]{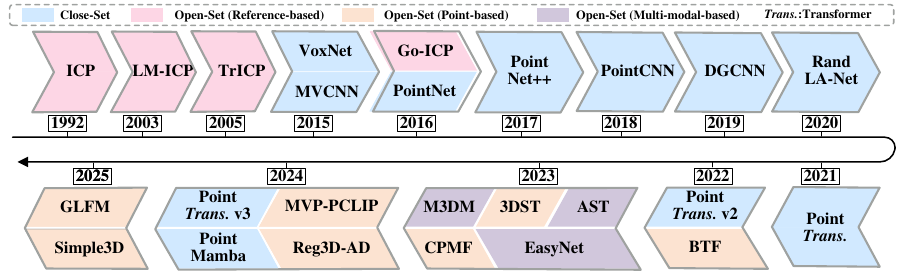}
\caption{\textbf{Timeline of representative methodologies in 3D DD.}}
\vspace{-3mm}
\label{fig:3DDD}
\end{figure*}

\section{3D Defect Detection}\label{sec:point_cloud_defect_detection}

Point cloud-based DD leverages its capacity to encapsulate the full 3D geometric and spatial attributes of an object, rendering it particularly adept for inspecting complex surfaces and irregular geometries. In contrast to image-based techniques, which are often impeded by occlusions, variable lighting conditions, or texture inconsistencies, point cloud methods deliver a more comprehensive and precise depiction of surface structures. This capability facilitates the accurate identification of defects---such as cracks, dents, or deformations---on intricate, curved surfaces.
The development of point cloud DD methods can be delineated into 
close-set detection and open-set detection, similar to the previous 2D defect detection methods. The representative methods are shown in Fig.~\ref{fig:3DDD} by chronological order.

\subsection{Close-Set 3D Defect Detection}
Close-set approaches leverage supervised deep learning to directly segment defects, typically formulating DD as a semantic segmentation task. They presuppose that all defect categories are known and represented within the training data, necessitating extensive labeled datasets. Based on their data representation, existing methods can be classified into two primary categories: point-based methods, which operate directly on raw point clouds, and methods employing alternative representations, such as voxelization and multi-view projection.

\subsubsection{Point-based methods}
Point-based methods process raw point clouds directly, addressing the challenge of learning permutation-invariant, point-wise features that effectively capture local contextual information. These methods are categorized by their feature extraction mechanisms: Multi-Layer Perceptron (MLP)-based, point convolution-based, graph convolution-based, and transformer-based approaches.

\noindent \textbf{MLPs-based Methods:}
PointNet \cite{Pointnet} pioneers the application of MLPs to extract features directly from point clouds. However, its architecture struggles to capture local structures induced by the metric space due to its global feature aggregation. To overcome this, PointNet++ \cite{PointNet++} employs hierarchical sampling via Farthest Point Sampling (FPS) and recursively applies PointNet to local neighborhoods, enhancing fine-grained pattern recognition and generalization to complex scenes. It introduces novel set learning layers to adaptively integrate multi-scale features. For large-scale point clouds, RandLA-Net \cite{RandLA-Net} expands the receptive field within a residual block framework. PointNeXt \cite{Pointnext} further refines scalability by integrating an inverted residual bottleneck and separable MLPs into PointNet++. To incorporate long-range context, 3P-RNN \cite{3P-RNN} proposes a two-step hierarchical Recurrent Neural Network (RNN) with one-stride multi-window pooling. Additionally, RepSurf \cite{RepSurf} enhances feature representation through novel triangular and umbrella surface constructs.

\noindent \textbf{Point Convolution-based Methods:}
Point convolution-based methods excel at preserving spatial information compared to MLP-based approaches. RSNet \cite{RS-Net} utilizes sequential 1$\times$1 convolution layers to generate independent feature representations for each point. PointwiseCNN \cite{pointwiseCNN} orders points (e.g., by XYZ coordinates or Morton curves) and applies 3D convolution to binned neighboring points within 3$\times$3$\times$3 kernel cells. PCCN \cite{PCCN} leverages MLPs to parameterize convolutional kernels across continuous vector spaces. However, direct convolution can discard shape information and introduce variance due to point ordering. PointCNN \cite{PointCNN} mitigates this by learning an $X$-transformation to weight and reorder input features into a canonical latent representation.

\noindent \textbf{Graph Convolution-based Methods:}
Graph convolution enhances local spatial aggregation and manages sparse point distributions effectively. LS-GCN \cite{LS-GCN} integrates spectral graph CNNs into the PointNet++ framework, recursively clustering spectral coordinates for feature aggregation. DGCNN \cite{DGCNN} introduces EdgeConv, dynamically constructing graphs per layer to capture long-range dependencies and local semantics. Superpoint Graph (SPG) \cite{SPG} models contextual relationships between object parts, enabling semantic segmentation of large-scale point clouds. To optimize computational efficiency, HDGCN \cite{HDGCN} employs depth-wise graph convolution followed by point-wise convolution. DeepGCNs \cite{DeepGCNs} incorporate residual connections to address vanishing gradient issues in deep architectures.

\noindent \textbf{Transformer-based Methods:}
Transformer-based methods leverage self-attention to capture long-range dependencies in point clouds. Point Transformer \cite{PointTransformer} designs tailored self-attention layers for point cloud tasks. Point Transformer V2 \cite{PointTransformer2} enhances this with grouped vector attention and position encoding multipliers. FPTransformer \cite{FPTransformer} adaptively weights local and global geometric connections for robust generalization. To reduce computational complexity, FastPointTransformer \cite{FastPointTransformer} employs voxel hashing, while Stratified-Transformer \cite{Stratified-Transformer} samples keys with a near-dense, far-sparse strategy. Point Transformer V3 \cite{PointTransformer3} replaces KNN-based neighbor search with serialized neighbor mapping, and PointMamba \cite{PointMamba} adapts a state space model for efficient global modeling.

\subsubsection{Other-formats-based methods}
The discrete, unstructured nature of point clouds poses computational challenges, prompting methods that convert them into structured formats. MVCNN \cite{MVCNN} renders point clouds into multi-view images processed by 2D CNNs, though optimal view selection remains complex. SqueezeSeg \cite{SqueezeSeg} parameterizes point clouds into range images using spherical coordinates, while SqueezeSegV2 \cite{SqueezeSegV2} adds a Context Aggregation Module for robustness. RangeViT \cite{RangeViT} applies Vision Transformers to range images. In constrast, 3D ShapeNets \cite{3D-ShapeNets} and VoxNet \cite{VoxNet} voxelize point clouds for 3D CNN segmentation. SEGCloud \cite{SEGCloud} refines voxel predictions with trilinear interpolation and Fully-Connected Conditional Random Fields for point-wise consistency. ScanComplete \cite{ScanComplete} uses a generative 3D CNN for scene completion, while OctNet \cite{OctNet} and SS-CNs \cite{SS-CNs} exploit sparsity via hierarchical partitioning and sparse convolution, respectively. Alternative formats like tangent images \cite{tangent} and lattices \cite{Splatnet} incur quantization losses due to projection errors.

\subsubsection{Remark}
The close-set detection methods have advanced point cloud segmentation significantly, yet their industrial deployment is constrained by fundamental limitations. They rely heavily on large, high-quality annotated datasets, which are costly and impractical to acquire for rare or diverse defects. Moreover, their specialization in predefined categories hampers generalization to novel anomalies, clashing with the open-ended nature of industrial inspection.

\begin{figure*}[h!]
\centering\includegraphics[width=\linewidth]{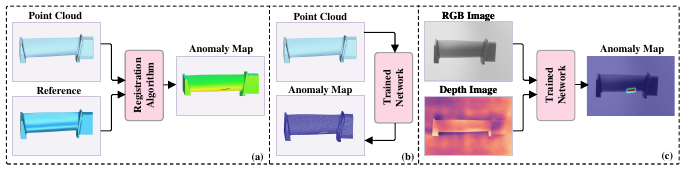}
\caption{\textbf{Representative schemes for 3D open-set DD.} (a) Reference-based anomaly detection leverages comparisons with  CAD models; (b) Point cloud-based anomaly detection operates directly on 3D geometry to identify irregularities; (c) Multi-modal-based anomaly detection integrates RGB images and depth images (or point clouds) to enhance robustness and accuracy.}

\vspace{-3mm}
\label{fig:Multi-modal}
\end{figure*}

\subsection{Open-Set 3D Defect Detection}
To address the limitations of closed-set detection methods in real-world industrial scenarios, open-set detection, particularly unsupervised anomaly detection, has emerged as a promising paradigm. Unlike supervised approaches that depend on annotated datasets and predefined defect classes, open-set methods identify anomalies without explicit supervision, enhancing adaptability to unseen defect types and reducing reliance on labor-intensive labeling. Current 3D detection techniques in this domain are broadly categorized into reference-based, point cloud-based, and multi-modal-based anomaly detection approaches.

\subsubsection{Reference-based Anomaly Detection}

Reference-based detection methods function by aligning the point cloud data of a test object with a reference digital model or point cloud. The methodology entails computing the geometric distance deviation of each point in the test point cloud relative to the reference, which serves as the anomaly score for defect identification, as shown in Fig. ~\ref{fig:Multi-modal} (a). The efficacy of these methods is predominantly contingent upon high-accuracy point cloud registration algorithms, tasked with determining the optimal transformation matrix $T^{*}$ that minimizes the alignment error between the test point cloud $X$ and the reference $Y$:

\begin{equation}
    T^{*} = \arg\min_{T \in SE(3)} \| Y - T X \| \label{F1}
\end{equation}
Point cloud registration is inherently an optimization challenge, and the solutions widely used for point cloud registration of industrial parts are divided into ICP-based, Probability-based, and Branch and Bound-based solutions.

\noindent\textbf{ICP-based Solution:}
ICP-based solution typically involves an iterative procedure of establishing correspondences between points in the test and reference point clouds and estimating the transformation matrix that minimizes the projection error of these correspondences. The Iterative Closest Point (ICP) algorithm~\cite{ICP} represents a pioneering contribution in this field. ICP operates by randomly selecting points in the test point cloud, identifying their nearest counterparts in the reference point cloud to form correspondences, and iteratively minimizing the Euclidean distance between these point pairs. However, its reliance on point-to-point distances introduces limitations, as these points may not exhibit exact one-to-one correspondence. To address this, extensions such as point-to-plane ICP~\cite{P2Plane_ISPRS, P2Plane_IJCV} minimize the orthogonal distance from points in one point cloud to the corresponding local plane in the other, enhancing correspondence accuracy. Further refinements, such as plane-to-plane correspondences~\cite{coarse, planetoplane}, bolster robustness.
% Addressing challenges in transformation estimation
A significant challenge in ICP-based methods is estimating the transformation matrix from extensive correspondence sets, risking convergence to local optima. To mitigate this, LM-ICP~\cite{LM_ICP} integrates an error correction mechanism during iterative optimization, while TrICP~\cite{TrICP} employs the Least Trimmed Square robust regression technique to enhance outlier resilience. Additionally, MVGR~\cite{MVGR} tackles uneven data density in industrial components by incorporating both distance and variance metrics of corresponding points.
% Evaluating the strengths and limitations
These ICP-based solutions are prized for their simplicity and efficiency, making them ideal for scenarios with clean, rigid point cloud data. However, their performance deteriorates markedly in the presence of noise, outliers, or non-rigid deformations.

\noindent\textbf{Probability-based Solution:}
% Outlining the probabilistic approach to registration
Probability-based methods conceptualize point cloud alignment as a probability density estimation problem. Techniques such as CPD~\cite{CPD} and GMMReg~\cite{GMMREG} fit Gaussian Mixture Models (GMMs) to the point clouds, maximizing likelihood or minimizing L2 distance between mixtures to estimate the optimal transformation. CPD models one point cloud as GMM centroids and the other as data points, whereas GMMReg focuses on mixture-to-mixture alignment.
% Enhancing robustness with advanced models
To address complex scenarios, ECMPR~\cite{ECMPR} introduces an Expectation Conditional Maximization (ECM) algorithm, accommodating general covariance matrices to surpass isotropic assumptions. JRMPC~\cite{JRMPC} extends this framework to multiple point clouds, employing batch and incremental EM algorithms to robustly estimate GMM parameters and transformations amidst noise and outliers. Furthermore, the Hybrid Mixture Model (HMM)~\cite{HMM} integrates a von-Mises-Fisher Mixture Model (FMM) for orientational uncertainty with a GMM for positional uncertainty, while LCGMM~\cite{LCGMM, PCLC} incorporates local consistency constraints to bolster registration robustness.
These methods excel in managing uncertainty and noise, offering superior stability over closed-form approaches. 

\noindent\textbf{Branch and Bound-based Solution:}
% Introducing BnB as a global optimization strategy
Branch-and-Bound (BnB)-based solutions systematically explore the solution space to secure the global optimum, partitioning it into subspaces and pruning suboptimal regions using computed bounds. The Box-and-Ball method~\cite{Box-and-ball} pioneers this approach, leveraging Lipschitz optimization and an octree structure for efficient search. BnBMIP~\cite{BnBMIP} reformulates registration as a mixed-integer programming problem, optimizing the consensus set for robust estimation.
% Integrating local and global techniques
Hybrid methods like Go-ICP~\cite{Go-ICP} embed local ICP iterations within a BnB framework, exploiting SE(3) geometry to derive novel error bounds, while GOGMA~\cite{GOGMA} achieves global optimality for Gaussian mixture alignment by minimizing L2 distance. Efficiency enhancements include stereographic projections~\cite{stereographic} for rapid match evaluation, ACM~\cite{ACM} for interval stabbing techniques, TR-DE~\cite{TR-DE} for parameter space decomposition, and TEAR~\cite{TEAR} for robust loss minimization via truncated residuals.
% Weighing optimality against efficiency
BnB methods ensure global optimality and robustness to initial conditions but incur significant computational costs, limiting their suitability for time-sensitive applications.

\subsubsection{Point Cloud-Based Anomaly Detection}
Point cloud-based methods typically aim to learn a compact representation of normal geometry and detect anomalies by identifying deviations in shape, topology, or spatial distribution, as shown in Fig. ~\ref{fig:Multi-modal} (b). These methods are particularly suited for tasks where color or texture is unreliable or unavailable, such as metallic surfaces or translucent materials.

\noindent\textbf{Teacher-Student Framework.} A common approach is the teacher-student framework, where a pre-trained model (teacher) guides an untrained model (student) to learn normal patterns. In point cloud anomaly detection, PointNet~\citep{Pointnet} is frequently adopted as the teacher. Early methods align features from a pre-trained PointNet teacher with an untrained student on normal data, using output discrepancies to detect anomalies. To enhance local geometric feature extraction, 3D-ST~\citep{3D-ST} employs self-supervised learning, reconstructing local receptive fields to improve the teacher's representation capacity.

\noindent\textbf{Reconstruction-Based Methods.} These methods reconstruct normal point cloud data, flagging anomalies via reconstruction errors. Masuda et al.~\citep{3DAutoencoder} pioneer this with a deep variational autoencoder, leveraging differences between input and reconstructed data. IMRNet~\citep{shape_anomaly} uses a point transformer backbone and introduces the Geometry-aware Point-cloud Sampling (GPS) module to partition point clouds into patches, applying a self-supervised masking strategy to boost reconstruction accuracy. R3D-AD~\citep{R3D-AD} leverages a diffusion-based generative model to obscure anomalous geometries during training, reconstructing normal patterns, and proposes Patch-Gen, a 3D anomaly simulation strategy, to bridge the training-testing domain gap. Uni-3DAD~\citep{Uni-3DAD} employs a GAN to reconstruct point clouds, targeting structural anomalies like missing parts.
MC3D-AD~\cite{MC3D-AD} introduces an adaptive geometry-aware masked attention module to guide mask attention.

\noindent\textbf{Prototype-Based Methods.} These methods~\cite{BTF,cpmf,real3d,Group3d,GLFM,Patch3D} have been widely adopted for their superior performance. These methods extract and store the features of normal data as normal prototypes, and obtain anomaly scores by calculating the distance between the test data features and the normal prototypes. BTF~\citep{BTF} uses handcrafted Fast Point Feature Histograms (FPFH). Reg3D-AD~\citep{real3d} combines raw coordinates with features from a pre-trained PointMAE~\citep{pointmae} model. ISMP~\citep{Lookinside} projects point clouds along multiple axes into pseudo-2D representations, supplementing global features. Group3AD~\citep{Group3d} and GLFM~\citep{GLFM} refine feature quality via self-supervised learning: Group3AD enhances cluster separation, while GLFM uses supervised segmentation on synthetic anomalies. Patch3D~\citep{Patch3D} segments point clouds into patches, creating region-specific memory banks to mitigate feature confusion. Simple3D~\cite{Simple3D}, only employing local descriptors, achieves real-time detection by feature aggregation.

\noindent\textbf{Multi-View Approaches.} Exploiting the robust capabilities of pre-trained image encoders, several methods transform point clouds into multi-view images to facilitate point-wise feature extraction~\cite{cpmf,PCLIP,PointAD}. Notably, CPMF~\cite{cpmf} pioneers this approach, while MVP-PCLIP~\cite{PCLIP} and PointAD~\cite{PointAD} further leverage advanced encoders to enhance feature representation.

\noindent\textbf{VLM-based Methods.} VLMs have emerged as powerful tools for point cloud anomaly detection due to their enhanced multimodal feature representation. Addressing the "cold start" challenge in industrial applications, MVP-PCLIP~\cite{PCLIP} and PointAD~\cite{PointAD} introduce zero-shot anomaly detection frameworks. These methods exploit the coherence between visual and textual features to identify anomalies without requiring training data. Specifically, MVP-PCLIP~\cite{PCLIP} employs a Key Layer Visual Prompt to refine visual features and designs Union and Specific text prompts to enrich textual descriptions. Additionally, PLANE~\cite{PLANE} synthesizes anomalies as training data, aligns features from point clouds and text encoders, and utilizes their consistency for effective anomaly detection.

\noindent\textbf{Others.} Beyond these paradigms, PO3AD~\cite{PO3AD} synthesizes anomalous samples and employs supervised learning to train a regression model for end-to-end anomaly score prediction. Meanwhile, advanced region growing techniques leverage local normals~\cite{lee2023new} and nearest neighbors~\cite{ye2025automatic} to directly segment anomalies in point clouds.

\subsubsection{Multi-modal-based Anomaly Detection}
Multi-modal-based anomaly detection integrates complementary modalities---such as point clouds, RGB images, and depth maps---to jointly capture geometric and appearance features, significantly improving robustness, as shown in Fig. ~\ref{fig:Multi-modal} (c). This fusion excels in identifying complex defects that single modalities struggle to detect.

\noindent\textbf{Multimodal Feature Fusion.} M3DM~\cite{M3DM} introduces unsupervised feature fusion by minimizing contrast loss across patches and establishing memory banks for RGB, 3D, and fused features, followed by decision-level fusion of detection outcomes. Subsequent research has refined feature fusion between RGB images and point clouds or depth maps. Shape-guided~\cite{shape_guided} proposes a Shape Expert to extract local point cloud features, aligning RGB feature vectors with 3D features to create dual memory banks for anomaly detection. To address overgeneralization in the teacher-student framework, AST~\cite{AST} adopts an asymmetric teacher model (conditional normalizing flow) and student models (CNNs), performing simultaneous knowledge distillation and feature comparison across RGB and depth images. EasyNet~\cite{EasyNet} mitigates interference from defect saliency disparities between RGB and depth images by synthesizing anomalies across modalities, using a reconstruction model to restore normal patterns and a segmentation model for anomaly detection. HOANG et al.~\cite{HOANG} enhance geometric feature reconstruction with a visual feature module incorporating non-local attention and graph convolutional networks, improving global-local feature interactions. Similarly, UCF~\cite{UCF} introduces discrepancy-guided fusion to assess embedding differences and their complementarities. Anomaly synthesis, as in EasyNet~\cite{EasyNet}, bolsters training robustness; for instance, 3DSR~\cite{3DSR} uses normal and synthetic anomalies with an autoencoder to reconstruct patterns, while 3DRÆM~\cite{3draem} targets industrial depth data synthesis.

\noindent\textbf{Crossmodal Reconstruction.} Crossmodal reconstruction enhances modality interaction by mutually reconstructing 2D and 3D information~\cite{Multimodal-rec}. CMDIAD~\cite{Incomplete} advances this with cross-modal distillation to handle incomplete modalities, while CPIR~\cite{CPIR} reconstructs 2D and 3D features, computing anomaly maps before and after reconstruction and fusing them via geometric mean.

\noindent\textbf{Teacher-Student Framework.} To counter underfitting due to modal discrepancies, CRD~\cite{CRD} extends single-branch distillation into a multi-branch framework, preserving anomaly sensitivity across modalities. LPFSTNet~\cite{LPFSTNet} integrates parameter-free head attention into the S-T framework for improved generalization, and MMRD~\cite{MMRD} incorporates reverse distillation. Conversely, 3D-ADNAS~\cite{3D-ADNAS} segments multi-level features into early, middle, and late stages, processing them sequentially through fusion modules to optimize strategy.

\noindent\textbf{Enriching Multi-Modal Spaces.} Additional modalities and architectures enrich multi-modal detection. 2M3DF~\cite{2M3DF} fuses colored multi-view images with point cloud features, 3D-MMFN~\cite{3D-MMFN} incorporates normal maps alongside RGB and point clouds, and MulSen-TripleAD~\cite{Mulsen} leverages infrared images to enhance 3D detection.

\noindent\textbf{Others.} Multi-modal anomaly detection grapples with domain shifts, few-shot/zero-shot scenarios, and noisy samples. LSFA~\cite{LSFA} bridges domain gaps by fine-tuning adaptors and learning task-oriented representations via intra-modal adaptation and cross-modal alignment. ITNM~\cite{ITNM} tackles incremental detection by blending new data features with memory banks and resampling coresets, using pixel positions to prevent misalignment. In few-shot settings, CLIP3D-AD~\cite{clip3d} colors point clouds with RGB and leverages CLIP for detection with minimal examples. For zero-shot, 3DzAL~\cite{3DzAL} uses contrastive learning with synthesized anomalies, and Zheng et al.~\cite{zheng} employ 2D/3D feature extraction with a voting mechanism. M3DM-NR~\cite{M3DM-NR} addresses noisy samples by using VLMs to extract 2D/3D features, correlating them with text to filter anomalies, and performing unsupervised detection on normal data.

\subsubsection{Remark}
% Analyzing strengths in controlled settings
Reference-based methods excel in controlled environments with high-fidelity reference models and strict adherence to design specifications, as often encountered in precision manufacturing~\cite{skin}. Their use of geometric deviation as an anomaly metric ensures precise defect localization. However, their reliance on high-precision registration falters under real-world conditions involving sensor noise, occlusion, surface variability, or non-rigid deformations. The assumption of an ideal reference model is frequently untenable amid tolerance-induced variations and intra-class variability, undermining generalization. While valuable in specific contexts, reference-based methods face significant challenges in robustness and adaptability.

Point cloud-based methods operate directly on 3D spatial data, offering a geometry-aware perspective for identifying structural deviations. However, these methods face two critical challenges that constrain their practical effectiveness. First, the absolute coordinate of points is inherently sensitive to sensor placement, calibration errors, and environmental disturbances, which typically damages accurate reconstruction of point clouds. Second, due to the sparse and irregular nature of point clouds, conventional deep learning backbones often struggle to extract meaningful and discriminative features. Compared with their 2D counterparts, point cloud-specific architectures generally exhibit limited expressive capacity, making it difficult to capture subtle geometric variations essential for detecting fine-grained defects.

Multi-modal-based methods attempt to integrate the appearance information from RGB images with geometric cues derived from depth maps. While these approaches provide complementary information, they inherit the inherent challenges of traditional 2D anomaly detection and further introduce additional modality-specific limitations. 
Modality mismatch often arises due to inherent differences in resolution and quality across sensors---for instance, RGB images typically provide high spatial resolution and rich texture, while depth maps are limited in resolution and may suffer from quantization or noise, making the fusion of such modalities suboptimal. Besides, modality alignment becomes increasingly difficult as more sensors are involved, such as incorporating infrared or hyperspectral data.

\subsection{Discussion}

Early close-set methods relying on supervised learning to identify predefined defect types, demonstrate strong performance only under controlled conditions. In contrast, open-set detection has emerged as a promising paradigm due to its capacity to identify diverse and previously unseen defect patterns. In open-set methods, recent advances, particularly in prototype-based and reconstruction-based methods, have shown promising potential in addressing the inherent challenges of industrial applications. Moreover, multimodal feature fusion frameworks further enhance detection robustness by leveraging complementary information across data modalities, thereby potentially improving reliability under varying material properties and illumination conditions. To accommodate extreme industrial scenarios, methodologies grounded in few-shot, zero-shot, and label-noise-resilient learning have also been proposed, enabling effective detection under conditions of severe data scarcity or uncertainty. Finally, incorporating physical priors and strengthening cross-domain feature generalization have become essential directions for building adaptable and reliable 3D DD systems.

\section{Future Directions}\label{sec:trend}

We have thoroughly evaluated the current landscape of 2D and 3D DD methods, encompassing both closed-set and open-set paradigms. In light of recent progress, several pivotal research directions emerge to overcome existing shortcomings and advance the field.

\noindent\textbf{Towards a Unified Defect Detection Framework.} Recent advancements in DD have shifted from closed-set methods~\citep{wang2024yolov10}, which are confined to recognizing known defect types, to open-set approaches~\citep{PatchCore} that excel in detecting novel anomalies. Within 3D DD, reference-based techniques utilizing standard CAD templates~\citep{ICP} enable precise deviation quantification. Real-world production environments frequently exhibit a dynamic evolution: initially mirroring open-set conditions with an absence of anomaly data, they progressively transition to closed-set scenarios as defect samples accrue. This underscores the necessity for a unified framework capable of seamlessly adapting to both contexts, thereby diminishing dependence on scenario-specific methodologies. Inspiration could be drawn from semi-supervised anomaly detection strategies~\citep{BiaS,BGAD} and cutting-edge unified models such as INP-Former++~\citep{INP-Former++}. Moreover, incorporating 3D defect quantification could substantially enhance downstream applications, including automated repair systems. Looking forward, we envision a versatile framework that operates effectively across closed-set and open-set domains, and potentially spans multiple modalities, thereby optimizing the defect detection pipeline.

\noindent\textbf{Towards Controllable Defect Generation.} The persistent scarcity of defect samples poses a significant challenge in DD, manifesting as limited instances of known defects and the absence of unseen defect types during training, aligning with few-shot~\citep{Hu_2024} and zero-shot~\citep{AnomalyAny} generation tasks, respectively. An effective framework must both augment known defects and synthesize novel ones, guided by expert knowledge embedded in textual descriptions from production standards~\citep{AnomalyAny,AnomalyVLM}. Future research should prioritize the development of controllable generation techniques, conditioned on defect attributes such as masks, textures, and material properties. Additionally, advancing generation models to produce defects consistent across modalities---such as 2D and 3D---would significantly enhance multimodal integration.

\noindent\textbf{Towards Configurable Defect Detection Systems.} Imaging quality, influenced by parameters including viewpoint, illumination, and sensor selection~\citep{M2AD}, critically impacts DD performance. The complexity of these factors necessitates expert oversight to balance quality and cost effectively. Configurable systems featuring automated parameter tuning could improve deployment efficiency, reduce expenses, and standardize imaging protocols industry-wide. With the growing prevalence of multimodal data---such as multi-view~\citep{real-iad}, multi-illumination~\citep{M2AD}, and RGB-D~\citep{MVTec3D} inputs---developing methods to harness these diverse sources is essential for achieving robust detection.

\noindent\textbf{Towards Large-Scale Real-World Datasets.} Foundational models, including MLLMs, rely on extensive, diverse datasets for optimal performance. However, current DD datasets, such as Real-IAD~\citep{real-iad} and M2AD~\citep{M2AD}, with fewer than one million samples, are considerably smaller than general computer vision datasets and are often limited to controlled laboratory conditions, reducing their real-world applicability. Collaborative initiatives uniting industry, government, and academia are vital to amass large-scale, real-world DD datasets, thereby improving model generalization and precision. Furthermore, employing generative models to create realistic, large-scale synthetic data offers substantial potential to supplement existing resources.

\noindent\textbf{Towards High-Resolution Defect Detection.} Real-world DD often requires identifying minute defects, such as 0.1 mm scratches on a 200 mm surface~\citep{VarAD,Simple3D}. Yet, existing methods, optimized for low-resolution inputs (e.g., 256$\times$256 pixels), fall short of these demands. High-resolution processing presents computational challenges, necessitating efficient architectures that integrate local and global contexts across scales. Future efforts should focus on scalable models that effectively reconcile computational constraints with high-resolution detection requirements.

\noindent\textbf{Towards High-Semantic Defect Detection.} Manufacturing defects encompass both structural and semantic categories. While structural defects are localized and detectable via local feature analysis, semantic defects---such as misassembled components~\citep{MVTecLOCO}---demand broader contextual comprehension. Recent developments leverage advanced architectures like ViTs~\citep{PNPT}, Mamba~\citep{VarAD,Mambaad}, and MLLMs~\citep{LogiCode,Anomaly-OV} to enhance semantic understanding. Continued advancement in this area is critical to addressing the diverse defect spectrum in manufacturing.

\noindent\textbf{Towards Enhanced Explainability.} Beyond anomaly detection, quantifying defects and tracing their origins are essential for downstream applications, such as repair and process optimization. Emerging techniques employing MLLMs for anomaly reasoning~\citep{MMAD,Anomaly-OV} demonstrate significant promise. Future research should integrate domain-specific expertise into these models, enhancing their explainability and yielding actionable insights to optimize manufacturing processes.

\section{Conclusion}\label{sec:conclusion}

This survey meticulously evaluates the trajectory of industrial DD methodologies, delineating their advancement from foundational closed-set techniques to contemporary open-set frameworks, spanning both 2D and 3D modalities. Our analysis underscores that open-set approaches are progressively establishing themselves as the predominant focus within this research sphere. We present an exhaustive taxonomy of these methodologies and extract insights poised to refine their real-world applications. It is our expectation that this survey will constitute a valuable reference for both researchers and practitioners alike, catalyzing further innovation within the realm of industrial DD.

% \bibliographystyle{elsarticle-num} 
% \bibliography{ref}

\end{document}